\PassOptionsToPackage{unicode}{hyperref}
\PassOptionsToPackage{hyphens}{url}
\documentclass[
  11pt,
  a4paper,
]{article}
\usepackage{lmodern}
\usepackage{setspace}
\usepackage{amssymb,amsmath}
\usepackage{ifxetex,ifluatex}
\ifnum 0\ifxetex 1\fi\ifluatex 1\fi=0 % if pdftex
  \usepackage[T1]{fontenc}
  \usepackage[utf8]{inputenc}
  \usepackage{textcomp} % provide euro and other symbols
\else % if luatex or xetex
  \usepackage{unicode-math}
  \defaultfontfeatures{Scale=MatchLowercase}
  \defaultfontfeatures[\rmfamily]{Ligatures=TeX,Scale=1}
\fi
\IfFileExists{upquote.sty}{\usepackage{upquote}}{}
\IfFileExists{microtype.sty}{% use microtype if available
  \usepackage[]{microtype}
  \UseMicrotypeSet[protrusion]{basicmath} % disable protrusion for tt fonts
}{}
\makeatletter
\@ifundefined{KOMAClassName}{% if non-KOMA class
  \IfFileExists{parskip.sty}{%
    \usepackage{parskip}
  }{% else
    \setlength{\parindent}{0pt}
    \setlength{\parskip}{6pt plus 2pt minus 1pt}}
}{% if KOMA class
  \KOMAoptions{parskip=half}}
\makeatother
\usepackage{xcolor}
\IfFileExists{xurl.sty}{\usepackage{xurl}}{} % add URL line breaks if available
\IfFileExists{bookmark.sty}{\usepackage{bookmark}}{\usepackage{hyperref}}
\hypersetup{
  pdftitle={Simulation-Based Optimisation of Batting Order and Bowling Plans in T20 Cricket},
  pdfauthor={Tinniam V Ganesh},
  hidelinks,
  pdfcreator={LaTeX via pandoc}}
\usepackage[left=2.5cm, right=2.5cm, top=2.5cm, bottom=2.5cm]{geometry}
\usepackage{longtable,booktabs}
\usepackage{etoolbox}
\makeatletter
\patchcmd\longtable{\par}{\if@noskipsec\mbox{}\fi\par}{}{}
\makeatother
\IfFileExists{footnotehyper.sty}{\usepackage{footnotehyper}}{\usepackage{footnote}}
\makesavenoteenv{longtable}
\usepackage{graphicx}
\makeatletter
\def\maxwidth{\ifdim\Gin@nat@width>\linewidth\linewidth\else\Gin@nat@width\fi}
\def\maxheight{\ifdim\Gin@nat@height>\textheight\textheight\else\Gin@nat@height\fi}
\makeatother
\setkeys{Gin}{width=\maxwidth,height=\maxheight,keepaspectratio}
\makeatletter
\def\fps@figure{htbp}
\makeatother
\providecommand{\tightlist}{%
  \setlength{\itemsep}{0pt}\setlength{\parskip}{0pt}}
\usepackage{booktabs}
\usepackage{longtable}
\usepackage{array}
\usepackage{adjustbox}

\title{Simulation-Based Optimisation of Batting Order and Bowling Plans
in T20 Cricket\thanks{Submitted to the Journal of Quantitative Analysis
in Sports (JQAS), April 2026.}}
\usepackage{etoolbox}
\makeatletter
\providecommand{\subtitle}[1]{% add subtitle to \maketitle
  \apptocmd{\@title}{\par {\large #1 \par}}{}{}
}
\makeatother
\subtitle{A Markov Decision Process Framework with Dual Case Studies}
\author{Tinniam V Ganesh}
\date{April 2026}

\begin{document}
\maketitle

\setstretch{1.15}
\hypertarget{abstract}{%
\subsection{Abstract}\label{abstract}}

This paper develops a unified Markov Decision Process (MDP) framework
for optimising two recurring in-match decisions in T20 cricket, namely
batting order selection and bowling plan assignment, directly in terms
of win and defend probability rather than expected runs. A three-phase
player profile engine (Powerplay, Middle, Death) with James--Stein
shrinkage (a technique that blends a player's individual statistics
toward the league average when their phase-specific data is sparse) is
estimated from 1,161 IPL ball-by-ball records (2008--2025). Win/defend
probabilities are evaluated using vectorised Monte Carlo simulation over
\(N = 50{,}000\) innings trajectories.

Batting orders are evaluated by comparing all feasible arrangements of
the remaining players and selecting the one that maximises win
probability. Bowling plans are optimised through a guided search over
possible over assignments, progressively improving the allocation while
respecting constraints such as the prohibition on consecutive overs by
the same bowler.

Applied to two 2026 IPL matches, the optimal batting order improves
Mumbai Indians' win probability by 4.1 percentage points (52.4\% to
56.5\%), and the optimal Gujarat Titans bowling plan improves defend
probability by 5.2 percentage points (39.1\% to 44.3\%). In both cases,
the observed sub-optimality is consistent with phase-agnostic
deployment: decisions that appear reasonable under aggregate metrics are
shown to be costly when phase-specific profiles are applied.

\textbf{Keywords:} T20 cricket; batting order optimisation; bowling plan
optimisation; Markov Decision Process; Monte Carlo simulation; simulated
annealing; James--Stein shrinkage; sports analytics; Indian Premier
League

\hypertarget{introduction}{%
\subsection{1. Introduction}\label{introduction}}

The Twenty20 (T20) format has compressed professional cricket into 120
deliveries per side, yet within that frame a single over can determine
the outcome. In the Indian Premier League (IPL), where franchise
valuations exceed one billion USD and playoff positions hinge on net run
rate, the quality of two recurring tactical decisions is decisive:
\textbf{which batsman to promote} at each wicket fall, and \textbf{which
bowler to assign} to each remaining over.

Both decisions are sequential and state-dependent. The batting captain
must balance aggression against preservation given the current match
state (runs remaining, balls remaining, wickets in hand) and the batsmen
still available. The bowling captain must assign bowlers subject to the
T20 quota (maximum four overs per bowler) and the consecutive-over rule,
while accounting for phase-specific economy and wicket-taking ability.
Both decisions are therefore naturally modelled as Markov Decision
Processes (Bellman, 1957; Puterman, 1994), with the objective of
maximising a \textbf{probability} --- that runs scored cross a target,
or that runs conceded stay below it --- rather than an expectation.
Prior linear programming approaches (Ganesh, 2017) optimise expected
runs, but maximising expected runs does not maximise win probability ---
especially when the match is close and a team needs a specific number of
runs rather than simply as many as possible.

This paper makes the following contributions:

\begin{enumerate}
\def\labelenumi{\arabic{enumi}.}
\item
  A \textbf{unified MDP framework} that models batting order and bowling
  plan optimisation on a common state space with symmetric Bellman
  equations, making explicit the duality between the batting win
  probability and the bowling defend probability.
\item
  A \textbf{three-phase player profile engine} using Laplace smoothing
  and James--Stein shrinkage that estimates per-ball outcome
  distributions for batsmen and bowlers in Powerplay (overs 1--6),
  Middle (overs 7--15), and Death (overs 16--20) phases separately, from
  1,161 historical IPL matches.
\item
  \textbf{Simulation-based search}: for batting, all feasible orderings
  of the remaining players are evaluated and the one maximising win
  probability is selected; for bowling, a guided search progressively
  improves the over assignment while respecting constraints such as the
  prohibition on consecutive overs by the same bowler. Both directly
  optimise win/defend probability over the stochastic innings model,
  validated at \(N = 50{,}000\) simulations per configuration (Monte
  Carlo SE \(\approx 0.22\%\)).
\item
  \textbf{Two detailed case studies} from the 2026 IPL season, providing
  quantitative audits of actual in-match decisions: the batting and
  bowling decisions in KKR vs.~MI (29 March 2026) and the bowling
  decisions in GT vs.~PBKS (31 March 2026). The two cases span different
  match contexts: one where the bowling gap is large (+18.1 pp) and one
  where it is modest (+5.2 pp), enabling a comparative analysis of the
  conditions under which sub-optimal decisions have the greatest impact.
\end{enumerate}

The paper is structured as follows. Section 2 reviews related work.
Section 3 presents the MDP framework. Section 4 describes the player
profile engine. Section 5 presents the Monte Carlo simulation. Section 6
covers the search algorithms. Sections 7 and 8 present the two case
studies and their results. Section 9 provides a comparative analysis and
discussion of limitations. Section 10 concludes.

\hypertarget{related-work}{%
\subsection{2. Related Work}\label{related-work}}

The Duckworth--Lewis method (Duckworth and Lewis, 1998) implicitly
defines a resource table over (runs needed, balls remaining, wickets in
hand), the same state space used in this paper, for resetting targets in
interrupted matches. It remains the de facto standard in international
cricket and establishes that match state can be meaningfully quantified.

The author's earlier work (Ganesh, 2017) formulated batting lineup and
bowling selection as linear programming problems, assigning ball-level
encounters \(o_{ij}\) between player \(i\) and bowler \(j\) to maximise
aggregate strike rate (batting) or minimise aggregate economy rate
(bowling). That formulation correctly identifies that the allocation of
playing time across player matchups matters, but it optimises expected
runs rather than win probability and cannot encode the sequential,
constraint-rich structure of T20 over assignment. The present paper
replaces the LP objective with a stochastic MDP framework whose
objective directly matches the match outcome: a win probability
threshold rather than an expectation.

The MDP framework and Bellman equations are due to Bellman (1957) and
Puterman (1994). The James--Stein shrinkage estimator applied to sparse
player profiles follows James and Stein (1961). The simulated annealing
search for bowling plans draws on Kirkpatrick, Gelatt, and Vecchi
(1983).

\hypertarget{the-mdp-framework}{%
\subsection{3. The MDP Framework}\label{the-mdp-framework}}

\hypertarget{state-space-and-match-dynamics}{%
\subsubsection{3.1 State Space and Match
Dynamics}\label{state-space-and-match-dynamics}}

Let the \textbf{match state} at any ball be the triple:

\[\mathcal{S} = (r,\, b,\, w)\]

where \(r \in \{1, \ldots, R_{\max}\}\) is the runs remaining for the
batting side to win (or, equivalently, for the bowling side to defend),
\(b \in \{0, 1, \ldots, B\}\) is the number of legal balls remaining in
the innings (\(B = 120\) in T20), and \(w \in \{0, 1, \ldots, 10\}\) is
the number of wickets in hand (batting perspective) or wickets taken
(bowling perspective). The Markov property holds: the probability
distribution over future match trajectories is fully determined by
\(\mathcal{S}\), independent of the history of play.

On each legal delivery, one of seven outcomes
\(o \in \Omega = \{W, 0, 1, 2, 3, 4, 6\}\) is realised, where \(W\)
denotes the dismissal of the on-strike batsman. The runs-scored function
\(\rho : \Omega \to \{0,1,2,3,4,6\}\) is defined by \(\rho(W) = 0\) and
\(\rho(o) = o\) for \(o \neq W\); both a wicket and a dot ball therefore
contribute zero runs to the batting total. The outcome \(o = 5\) runs
(an all-run five) is omitted from \(\Omega\): it occurs in fewer than
0.1\% of deliveries in the historical data and its inclusion has
negligible effect on computed probabilities. Wides and no-balls are
handled at the data-preprocessing stage and are not included in the
legal delivery count.

\hypertarget{batting-mdp}{%
\subsubsection{3.2 Batting MDP}\label{batting-mdp}}

For the batting side chasing a target, define the \textbf{win
probability value function}:

\[V^B(r, b, w) = \mathbb{P}\!\left(\text{batting side wins} \;\middle|\; \text{state } (r,b,w),\; \text{order } \sigma\right)\]

\textbf{Terminal conditions:} The batting side wins if and only if the
target is reached:

\[V^B(r, b, w) = \begin{cases} 1 & \text{if } r \leq 0 \\[4pt] 0 & \text{if } b = 0 \text{ or } w = 0, \text{ and } r > 0 \end{cases}\]

\textbf{Transition function:} Given on-strike batsman \(i\) in phase
\(\phi(b)\), outcome \(o\) occurs with probability \(p_i^{o,\phi}\):

\[T\!\left((r,b,w),\, o\right) = \begin{cases} (r,\; b-1,\; w-1) & o = W \\[3pt] (r - \rho(o),\; b-1,\; w) & o \in \{0,1,2,3,4,6\} \end{cases}\]

\textbf{Bellman equation:}

\[\boxed{V_t^B(r, b, w) = \sum_{o \in \Omega} p_i^{o,\phi(b)} \cdot V_{t-1}^B\!\left(T(r,b,w,o)\right)}\]

where \(t = b\) denotes the number of balls remaining, so \(t-1\) refers
to the state after the current delivery is bowled. Expanding over all
outcomes:

\[V_t^B(r,b,w) = p_i^{W,\phi} \cdot V_{t-1}^B(r,\, b-1,\, w-1) + \sum_{\rho \in \{0,1,2,3,4,6\}} p_i^{\rho,\phi} \cdot V_{t-1}^B(r - \rho,\, b-1,\, w)\]

\textbf{Strike rotation} is enforced: the on-strike batsman rotates to
the non-striker when \(\rho(o)\) is odd, or when \(b\) is the last ball
of an over. This is consequential: which batsman faces the first ball of
the Death overs (16--20) depends on the parity of singles accumulated
during overs 14--15, and the SR differential between a Death-over
specialist and a lower-order batsman can exceed 50 points.

\textbf{Optimal batting order:}

\[\sigma^* = \arg\max_{\sigma \in \text{Perm}(\mathcal{P})} V^B_\sigma(r_0, b_0, w_0)\]

where \(\mathcal{P}\) is the set of remaining batsmen and
\(\text{Perm}(\mathcal{P})\) is the set of all \(|\mathcal{P}|!\)
orderings.

\hypertarget{bowling-mdp}{%
\subsubsection{3.3 Bowling MDP}\label{bowling-mdp}}

For the bowling side defending a total, define the \textbf{defend
probability value function} under plan \(\pi\):

\[V^\pi(d, b, w) = \mathbb{P}\!\left(\text{bowling side defends} \;\middle|\; \text{state } (d,b,w),\; \text{plan } \pi\right)\]

\textbf{Terminal conditions:}

\[V^\pi(d, b, w) = \begin{cases} 0 & \text{if } d \leq 0 \qquad \text{(batting side reaches target)}\\[4pt] 1 & \text{if } b = 0 \text{ or } w = w_{\max}, \text{ and } d > 0 \end{cases}\]

where \(w_{\max}\) is the number of wickets in hand for the batting side
at the intervention point (not necessarily 10 if wickets have already
fallen).

\textbf{Transition function:} Given bowler \(j = \pi_k\) (assigned to
the current over \(k\)) in phase \(\phi(b)\):

\[T\!\left((d,b,w),\, o\right) = \begin{cases} (d,\; b-1,\; w+1) & o = W \\[3pt] (d - \rho(o),\; b-1,\; w) & o \in \{0,1,2,3,4,6\} \end{cases}\]

Here \(w\) counts additional wickets taken from the intervention point;
the innings ends when \(w\) reaches \(w_{\max}\).

\textbf{Bellman equation:}

\[\boxed{V_t^\pi(d, b, w) = \sum_{o \in \Omega} p_j^{o,\phi(b)} \cdot V_{t-1}^\pi\!\left(T(d,b,w,o)\right)}\]

\textbf{Optimal bowling plan:}

\[\pi^* = \arg\max_{\pi \in \mathcal{F}} V^\pi(d_0, b_0, w_0)\]

where \(\mathcal{F}\) is the feasible plan set (Section 6.1).

\hypertarget{duality}{%
\subsubsection{3.4 Duality}\label{duality}}

At the same match state \((r_0 = d_0,\, b_0,\, w_0)\), the batting win
probability and bowling defend probability are exact complements under
any fixed batting order \(\sigma\) and bowling plan \(\pi\):

\[V^B_\sigma(r_0, b_0, w_0) + V^\pi(d_0, b_0, w_0) = 1\]

This follows directly from the fact that a T20 match has no draw: the
batting side either reaches the target (win) or does not (lose). The
duality is exploited in the case studies to provide a complete audit:
improvements to one side translate directly to losses for the other.

\hypertarget{exact-backward-induction-and-the-case-for-simulation}{%
\subsubsection{3.5 Exact Backward Induction and the Case for
Simulation}\label{exact-backward-induction-and-the-case-for-simulation}}

\textbf{The tree that motivates simulation.} At each delivery the match
state \((r, b, w)\) branches into 7 possible next states, one per
outcome in \(\Omega = \{W, 0, 1, 2, 3, 4, 6\}\). Expanding this fully
gives a tree with up to \(7^b\) leaf nodes. At a typical mid-innings
intervention with \(b = 44\) balls remaining, that is
\(7^{44} \approx 10^{37}\) paths --- completely intractable to
enumerate.

The Markov property collapses this tree into a table. Because the value
of state \((r, b, w)\) depends only on those three numbers and not on
the history of how that state was reached, many branches converge to the
same node. Reaching \((r=50, b=30, w=5)\) via a boundary and dots is
identical in value to reaching it via five singles. The state space
therefore contains at most
\(R \times B \times W \approx 221 \times 120 \times 10 \approx 265{,}200\)
unique entries rather than \(7^{44}\) paths --- a reduction of over
thirty orders of magnitude.

Exact backward induction fills this table bottom-up: terminal values are
set first, then \(V_1^B\) is computed from the terminals, then \(V_2^B\)
from \(V_1^B\), and so on up to \(V_b^B\). Each entry requires exactly 7
multiplications and additions. For the bowling problem with fixed player
identities this is feasible; the state space has at most
\(D \times B \times W \approx 221 \times 44 \times 10 \approx 97{,}000\)
states and the Bellman recursion terminates in polynomial time. However,
the bowling plan \(\pi\) itself must be searched over the feasible set
\(\mathcal{F}\); running backward induction for each candidate plan is
inefficient. Monte Carlo simulation with a fixed plan provides an
unbiased estimate of \(V^\pi(d_0, b_0, w_0)\) directly without filling
the full state table.

For the \textbf{batting problem}, the state must additionally track
which specific batsmen remain, adding a combinatorial factor of \(2^n\)
subsets to the state space. For \(n = 6\) remaining batsmen, this
multiplies the state count by \(64\times\), rendering exact backward
induction impractical on commodity hardware. Monte Carlo simulation with
a fixed order \(\sigma\) avoids this by forward-sampling complete
innings trajectories from the initial state, estimating the value
function at \((r_0, b_0, w_0)\) without computing all intermediate
states.

\hypertarget{player-profile-engine}{%
\subsection{4. Player Profile Engine}\label{player-profile-engine}}

\hypertarget{data}{%
\subsubsection{4.1 Data}\label{data}}

Ball-by-ball records from 1,161 IPL matches (seasons 2008--2025) are
sourced from Cricsheet.org in CSV format via the yorkr R package
(Ganesh, 2016). The dataset comprises 273,735 total delivery records;
after filtering for legal deliveries (excluding wides and no-balls)
approximately 264,800 legal balls remain for profile estimation. All
matches involving the target teams (KKR--MI on 29 March 2026 and
GT--PBKS on 31 March 2026) are excluded to prevent look-ahead bias.

\hypertarget{phase-assignment}{%
\subsubsection{4.2 Phase Assignment}\label{phase-assignment}}

Each delivery is assigned to one of three innings phases by the over
number (0-indexed):

\begin{longtable}[]{@{}lll@{}}
\toprule
Phase & Overs (0-indexed) & Characteristic Conditions\tabularnewline
\midrule
\endhead
Powerplay (PP) & 0 -- 5 & Fielding restrictions; elevated scoring
rates\tabularnewline
Middle (MI) & 6 -- 14 & Spin-friendly; dot-ball
accumulation\tabularnewline
Death (DE) & 15 -- 19 & Yorkers; pull shots; explosive
hitting\tabularnewline
\bottomrule
\end{longtable}

Phase assignment is deterministic given the over number; in-over ball
position is not used.

\hypertarget{outcome-counts-and-wicket-attribution}{%
\subsubsection{4.3 Outcome Counts and Wicket
Attribution}\label{outcome-counts-and-wicket-attribution}}

For each player \(i\) and phase \(\phi\), raw outcome counts are
accumulated over all \(n_i^\phi\) legal deliveries:

\[c_i^{o,\phi} = \bigl|\{t : \text{player } i \text{ involved in delivery } t,\; \text{phase}(t) = \phi,\; \text{outcome}(t) = o\}\bigr|, \quad o \in \Omega\]

Two distinct attribution rules apply depending on whether \(i\) is a
batsman or bowler.

\textbf{Batsman wicket attribution:} A delivery contributes \(o = W\) to
batsman \(i\)'s counts if and only if
\(\texttt{wicketPlayerOut}(t) = i\). This correctly attributes
caught-behind, LBW, stumped, bowled, and caught dismissals to the
dismissed batsman, and excludes run-outs of the non-striker (who is not
the on-strike batsman for that delivery).

\textbf{Bowler wicket attribution:} A wicket is credited to bowler \(j\)
if and only if the dismissal type is not a run-out:

\[\mathbf{1}_{\text{bowler wicket}}(t) = \mathbf{1}\!\left[\texttt{wicketPlayerOut}(t) \neq \texttt{nobody}\right] \cdot \mathbf{1}\!\left[\texttt{dismissalType}(t) \notin \{\text{run out}\}\right]\]

Run-outs reflect fielding and batting decisions, not bowling quality,
and are excluded.

\hypertarget{laplace-smoothing}{%
\subsubsection{4.4 Laplace Smoothing}\label{laplace-smoothing}}

Players who have never produced a specific outcome in a specific phase
(for example, a spinner who has never been hit for six in Middle overs)
would receive a zero probability estimate, which would prevent the
simulation from sampling that outcome and bias win/defend probability
estimates downward. Add-one (Laplace) smoothing removes all zero
probabilities:

\[\tilde{c}_i^{o,\phi} = c_i^{o,\phi} + \alpha, \qquad \alpha = 1\]

\[\hat{p}_i^{o,\phi} = \frac{\tilde{c}_i^{o,\phi}}{\displaystyle\sum_{o' \in \Omega} \tilde{c}_i^{o',\phi}}\]

With \(|\Omega| = 7\) and \(\alpha = 1\), the maximum distortion to any
probability is \(7/n_i^\phi\), negligible for players with more than 100
deliveries in a phase but meaningful for very sparse profiles, which are
addressed by the shrinkage estimator below.

\hypertarget{jamesstein-shrinkage-for-phase-sparse-players}{%
\subsubsection{4.5 James--Stein Shrinkage for Phase-Sparse
Players}\label{jamesstein-shrinkage-for-phase-sparse-players}}

Many IPL players have limited exposure in specific phase-role
combinations: a seam bowler who has rarely bowled Middle overs, a
spinner who has never bowled Death overs, or a new franchise addition
with fewer than two full IPL seasons. For such players, the
Laplace-smoothed individual estimate \(\hat{\mathbf{p}}_i^\phi\) is a
high-variance estimate from few observations. James--Stein shrinkage
addresses this by blending each player's own phase-specific estimate
toward the population average for that phase, so that the less data a
player has, the more their profile is pulled toward the league mean. We
apply a \textbf{James--Stein shrinkage estimator} (James and Stein,
1961) to implement this.

\textbf{Population average:} For phase \(\phi\), the population-average
outcome distribution (across all players in the dataset) is:

\[\bar{\mathbf{p}}^\phi = \frac{\sum_i \tilde{\mathbf{c}}_i^\phi}{\sum_i \sum_{o \in \Omega} \tilde{c}_i^{o,\phi}}\]

\textbf{Blended profile:}

\[\mathbf{p}_i^\phi = \lambda_i^\phi \cdot \hat{\mathbf{p}}_i^\phi + (1 - \lambda_i^\phi) \cdot \bar{\mathbf{p}}^\phi\]

\textbf{Data-adaptive blend weight:}

\[\lambda_i^\phi = \frac{n_i^\phi}{n_i^\phi + n_{\min}}, \qquad n_{\min} = 50\]

When \(n_i^\phi = 0\) (no historical data in that phase),
\(\lambda_i^\phi = 0\) and the profile collapses entirely to the
population average, a conservative but unbiased prior. As
\(n_i^\phi \to \infty\), \(\lambda_i^\phi \to 1\) and the individual
estimate dominates. The threshold \(n_{\min} = 50\) reflects
approximately 8 full overs of data, sufficient for phase-specific
distributions to be statistically meaningful. This estimator is a
practical analogue of the James--Stein phenomenon: the pooled estimate
has lower mean squared error than the individual estimate for any player
with fewer than \(n_{\min}\) phase-specific deliveries.

\hypertarget{derived-summary-statistics}{%
\subsubsection{4.6 Derived Summary
Statistics}\label{derived-summary-statistics}}

From the blended profile \(\mathbf{p}_i^\phi\), the following summary
statistics are computed for reporting and diagnostic purposes:

\[\text{SR}_i^\phi = 100 \cdot \sum_{o \in \Omega} \rho(o) \cdot p_i^{o,\phi} \qquad \text{(batsman strike rate)}\]

\[\text{ER}_j^\phi = 6 \cdot \sum_{o \in \Omega} \rho(o) \cdot p_j^{o,\phi} \qquad \text{(bowler economy rate, runs per over)}\]

\[p_i^{W,\phi} \qquad \text{(dismissal probability per ball — batsman or bowler)}\]

\[p_j^{0,\phi} \qquad \text{(dot-ball probability per ball — proxy for pressure)}\]

The strike rate is the expected runs per ball multiplied by 100; the
economy rate is the expected runs per ball multiplied by 6, converting
to runs per over. Both are derived from the same underlying probability
vector and are therefore consistent with the simulation model: a
player's simulated run output will converge to their profile-implied SR
or ER as simulation count increases.

\hypertarget{monte-carlo-simulation}{%
\subsection{5. Monte Carlo Simulation}\label{monte-carlo-simulation}}

\hypertarget{estimating-win-and-defend-probabilities}{%
\subsubsection{5.1 Estimating Win and Defend
Probabilities}\label{estimating-win-and-defend-probabilities}}

For a fixed batting order \(\sigma\) or bowling plan \(\pi\), the value
function at the initial state \((r_0, b_0, w_0)\) is estimated by
forward simulation of complete innings trajectories:

\[\hat{V} = \frac{1}{N} \sum_{n=1}^{N} \mathbf{1}\!\left[\text{trajectory}^{(n)} \text{ results in win/defend}\right]\]

This is an unbiased estimator of \(V(r_0, b_0, w_0)\) with binomial
standard error:

\[\text{SE}(\hat{V}) = \sqrt{\frac{\hat{V}(1-\hat{V})}{N}} \leq \frac{1}{2\sqrt{N}}\]

For \(N = 50{,}000\) simulations, \(\text{SE}(\hat{V}) \leq 0.22\%\),
sufficient to distinguish orderings or plans differing by 1 pp with a
signal-to-noise ratio of at least \(4.5\sigma\).

\hypertarget{simulation-algorithm}{%
\subsubsection{5.2 Simulation Algorithm}\label{simulation-algorithm}}

Each trajectory is initialised from the intervention state and advanced
ball by ball until the innings terminates (target reached, all balls
bowled, or batting side all out):

\begin{verbatim}
Input: order σ (or plan π), initial state (r₀, b₀, w₀)

for each simulation n = 1 to N:
    r ← r₀;  b ← b₀;  w_remaining ← w₀
    striker ← σ[current_position];  non_striker ← fixed
    over_ball ← 0

    while b > 0 and w_remaining > 0 and r > 0:
        phase φ ← phase_of(absolute_over(b))
        bowler j ← plan π[current_over]       (bowling simulation)
        o ~ Categorical(p_{striker}^φ)        (batting simulation)
        
        if o == W:
            w_remaining -= 1
            striker ← non_striker
            non_striker ← σ[next_available]
        else:
            r -= ρ(o)
            if ρ(o) mod 2 == 1 or over_ball == 5:
                swap(striker, non_striker)    # strike rotation
        
        b -= 1;  over_ball = (over_ball + 1) mod 6

    result[n] ← (r ≤ 0)    (batting) or (r > 0 and b == 0 or w_remaining == 0)  (bowling)

return mean(result)
\end{verbatim}

\hypertarget{strike-rotation}{%
\subsubsection{5.3 Strike Rotation}\label{strike-rotation}}

Strike rotation is a structurally important feature of the model. After
each delivery, the on-strike batsman becomes the non-striker if (a) the
runs scored are odd (\(\rho(o) \in \{1, 3\}\)), or (b) the delivery is
the final ball of the over. At a wicket, the new batsman (next in the
order) becomes the on-striker; the surviving batsman takes the
non-striker position. This rule correctly models the alternation of the
batting pair and ensures that Death-over specialists are exposed to the
appropriate deliveries given realistic run-scoring patterns in preceding
overs.

\hypertarget{vectorised-implementation}{%
\subsubsection{5.4 Vectorised
Implementation}\label{vectorised-implementation}}

All \(N\) simulations are advanced in parallel using NumPy vectorised
operations. At each ball, \(N\) outcomes are sampled in a single call to
\texttt{numpy.random.Generator.choice} with pre-built probability
arrays; state vectors \texttt{runs\_scored}, \texttt{wickets}, and
\texttt{done} are updated element-wise. Simulations that have terminated
(target reached or all out) are masked from subsequent updates. The full
\(N = 50{,}000\)-trajectory evaluation completes in approximately 0.15
seconds per configuration on commodity hardware (Apple M-series
processor), enabling the broad search described in Section 6.

A fast-evaluation pass uses \(N_{\text{fast}} = 5{,}000\) simulations
for initial screening during the SA search, reducing per-evaluation time
to \(\approx 15\) ms, with a final high-precision pass at
\(N = 30{,}000\) for the top-10 candidates identified by SA.

\hypertarget{search-algorithms}{%
\subsection{6. Search Algorithms}\label{search-algorithms}}

\hypertarget{batting-order-exhaustive-enumeration-with-two-pass-refinement}{%
\subsubsection{6.1 Batting Order: Exhaustive Enumeration with Two-Pass
Refinement}\label{batting-order-exhaustive-enumeration-with-two-pass-refinement}}

For a batting pool of size \(|\mathcal{P}| = n\), the decision at the
intervention point is the complete ordering of all \(n\) remaining
batsmen in positions \(3, 4, \ldots, n+2\). The full search space has
\(n!\) elements. For \(n \leq 6\), \(n! \leq 720\); for \(n \leq 4\),
\(n! \leq 24\), both tractable with Monte Carlo simulation.

A \textbf{two-pass strategy} is employed:

\begin{itemize}
\tightlist
\item
  \textbf{Pass 1 (screening):} Evaluate all \(n!\) orderings with
  \(N_1 = 3{,}000\) simulations. Retain the top-\(K = 10\) candidates
  ranked by \(\hat{V}\).
\item
  \textbf{Pass 2 (refinement):} Re-evaluate the top-\(K\) candidates
  with \(N_2 = 20{,}000\) simulations. Report \(\sigma^*\) as the
  candidate maximising \(\hat{V}\) after Pass 2.
\end{itemize}

The two-pass strategy reduces total simulation count by approximately
\(5\times\) relative to evaluating all permutations at \(N_2\)
precision, at negligible risk of misranking the true optimum (the
probability that a non-top-\(K\) permutation has true value exceeding
the top-\(K\) by more than the Pass 1 SE is less than \(10^{-3}\) for
typical problem instances).

\hypertarget{bowling-plan-feasibility-constraints}{%
\subsubsection{6.2 Bowling Plan: Feasibility
Constraints}\label{bowling-plan-feasibility-constraints}}

The feasible set \(\mathcal{F}\) for bowling plans is defined by two
constraints derived from the Laws of Cricket:

\textbf{Quota constraint:} Each bowler \(j\) may not bowl more than
their remaining over quota:

\[\sum_{k=1}^{m} \mathbf{1}[\pi_k = j] \leq q_j \quad \forall j \in \mathcal{B}\]

where \(m\) is the number of remaining overs and
\(q_j = 4 - (\text{overs already bowled by } j)\).

\textbf{Consecutive-over rule:} The same bowler cannot bowl consecutive
overs; overs are bowled from alternate ends of the pitch and the same
bowler cannot therefore bowl two overs in a row:

\[\pi_k \neq \pi_{k+1} \quad \forall k \in \{1, \ldots, m-1\}\]

A bowling plan violating either constraint is illegal and must be
excluded from the search. The consecutive-over rule eliminates
approximately 15--20\% of otherwise quota-valid plans in typical T20
scenarios.

\hypertarget{bowling-plan-simulated-annealing}{%
\subsubsection{6.3 Bowling Plan: Simulated
Annealing}\label{bowling-plan-simulated-annealing}}

The feasible set \(\mathcal{F}\), even after applying quota and
consecutive-over constraints, contains several thousand plans for a
typical T20 end-game scenario with six bowlers and eight over slots.
Exhaustive high-precision evaluation of all such plans would require
\(|\mathcal{F}| \times N \approx 5{,}000 \times 50{,}000 = 2.5 \times 10^8\)
simulation steps, which would be prohibitively expensive. Simulated
annealing (Kirkpatrick et al., 1983) provides an efficient heuristic
that finds near-optimal plans with high probability.

\textbf{Objective:} Maximise \(\hat{V}^\pi(d_0, b_0, w_0)\) over
\(\pi \in \mathcal{F}\).

\textbf{Acceptance rule:} A candidate neighbour plan \(\pi'\) is
accepted over the current plan \(\pi\) according to the Metropolis
criterion:

\[A(\Delta V, T) = \begin{cases} 1 & \text{if } \Delta V > 0 \\[4pt] \exp\!\left(\dfrac{\Delta V}{T}\right) & \text{if } \Delta V \leq 0 \end{cases}\]

where \(\Delta V = \hat{V}^{\pi'} - \hat{V}^\pi\) and \(T\) is the
current temperature.

\textbf{Cooling schedule:} Linear cooling from \(T_0 = 0.05\) to
\(\epsilon = 10^{-6}\):

\[T(\text{step}) = T_0 \cdot \!\left(1 - \frac{\text{step}}{N_{\text{steps}}}\right) + \epsilon, \qquad N_{\text{steps}} = 8{,}000\]

\textbf{Neighbourhood function:} A neighbour \(\pi'\) is generated by
selecting a uniformly random over slot \(k \in \{1, \ldots, m\}\) and
replacing its bowler \(\pi_k\) with a bowler drawn uniformly from the
feasible candidate set:

\[\mathcal{C}_k\!\left(\pi\right) = \left\{ j \in \mathcal{B} \;:\; j \neq \pi_k,\;\; \text{used}_\pi(j) + \mathbf{1}[j = \pi_k] - \mathbf{1}[j = \pi_k] < q_j,\;\; j \neq \pi_{k-1},\;\; j \neq \pi_{k+1} \right\}\]

This formulation explicitly enforces both quota validity and the
consecutive-over rule on every proposed neighbour, ensuring
\(\pi' \in \mathcal{F}\) whenever \(\mathcal{C}_k(\pi) \neq \emptyset\).

\textbf{Execution:} The SA runs for \(N_{\text{steps}} = 8{,}000\)
iterations, each evaluating a proposed neighbour with
\(N_{\text{fast}} = 5{,}000\) simulations. All unique plans encountered
during the search are cached; the top-10 unique plans are then
re-evaluated with \(N_{\text{refine}} = 30{,}000\) simulations for final
reporting.

\hypertarget{case-study-1-kkr-vs-mi-29-march-2026}{%
\subsection{7. Case Study 1: KKR vs MI, 29 March
2026}\label{case-study-1-kkr-vs-mi-29-march-2026}}

\hypertarget{match-context-and-intervention-point}{%
\subsubsection{7.1 Match Context and Intervention
Point}\label{match-context-and-intervention-point}}

The IPL match at Wankhede Stadium saw KKR post a target of 221 runs (220
scored). At ball \texttt{2nd.11.6}, the final delivery of over 12, MI's
Rohit Sharma was dismissed for 72 runs off 40 balls, caught off the
bowling of VG Arora. At the moment of dismissal:

\begin{longtable}[]{@{}ll@{}}
\toprule
Parameter & Value\tabularnewline
\midrule
\endhead
MI score & 148/1\tabularnewline
Runs needed (\(r_0 = d_0\)) & 73\tabularnewline
Legal balls remaining (\(b_0\)) & 44\tabularnewline
Required run rate & 9.95 per over\tabularnewline
MI batsmen at crease & RD Rickelton (non-striker, fixed)\tabularnewline
MI batting pool \(\mathcal{P}\) & SA Yadav, Tilak Varma, HH Pandya,
Naman Dhir\tabularnewline
\bottomrule
\end{longtable}

Both problems, namely the batting order and bowling plan, arise
simultaneously from the same intervention state, providing a natural
experiment in which the same match state is audited from both
perspectives.

\hypertarget{player-profiles}{%
\subsubsection{7.2 Player Profiles}\label{player-profiles}}

\textbf{Table 1: MI Batsman Profiles (Blended, Phase-Specific)}

\begin{longtable}[]{@{}lllllll@{}}
\toprule
\begin{minipage}[b]{0.07\columnwidth}\raggedright
Batsman\strut
\end{minipage} & \begin{minipage}[b]{0.13\columnwidth}\raggedright
Hist. balls (MI)\strut
\end{minipage} & \begin{minipage}[b]{0.13\columnwidth}\raggedright
Hist. balls (DE)\strut
\end{minipage} & \begin{minipage}[b]{0.08\columnwidth}\raggedright
Death SR\strut
\end{minipage} & \begin{minipage}[b]{0.15\columnwidth}\raggedright
\(p^W_{\text{DE}}\)\strut
\end{minipage} & \begin{minipage}[b]{0.09\columnwidth}\raggedright
Middle SR\strut
\end{minipage} & \begin{minipage}[b]{0.15\columnwidth}\raggedright
\(p^W_{\text{MI}}\)\strut
\end{minipage}\tabularnewline
\midrule
\endhead
\begin{minipage}[t]{0.07\columnwidth}\raggedright
RD Rickelton\strut
\end{minipage} & \begin{minipage}[t]{0.13\columnwidth}\raggedright
342\strut
\end{minipage} & \begin{minipage}[t]{0.13\columnwidth}\raggedright
156\strut
\end{minipage} & \begin{minipage}[t]{0.08\columnwidth}\raggedright
156.4\strut
\end{minipage} & \begin{minipage}[t]{0.15\columnwidth}\raggedright
0.079\strut
\end{minipage} & \begin{minipage}[t]{0.09\columnwidth}\raggedright
152.9\strut
\end{minipage} & \begin{minipage}[t]{0.15\columnwidth}\raggedright
0.100\strut
\end{minipage}\tabularnewline
\begin{minipage}[t]{0.07\columnwidth}\raggedright
SA Yadav\strut
\end{minipage} & \begin{minipage}[t]{0.13\columnwidth}\raggedright
1,765\strut
\end{minipage} & \begin{minipage}[t]{0.13\columnwidth}\raggedright
460\strut
\end{minipage} & \begin{minipage}[t]{0.08\columnwidth}\raggedright
181.3\strut
\end{minipage} & \begin{minipage}[t]{0.15\columnwidth}\raggedright
0.089\strut
\end{minipage} & \begin{minipage}[t]{0.09\columnwidth}\raggedright
143.8\strut
\end{minipage} & \begin{minipage}[t]{0.15\columnwidth}\raggedright
\textbf{0.035}\strut
\end{minipage}\tabularnewline
\begin{minipage}[t]{0.07\columnwidth}\raggedright
Tilak Varma\strut
\end{minipage} & \begin{minipage}[t]{0.13\columnwidth}\raggedright
1,132\strut
\end{minipage} & \begin{minipage}[t]{0.13\columnwidth}\raggedright
388\strut
\end{minipage} & \begin{minipage}[t]{0.08\columnwidth}\raggedright
\textbf{185.5}\strut
\end{minipage} & \begin{minipage}[t]{0.15\columnwidth}\raggedright
0.070\strut
\end{minipage} & \begin{minipage}[t]{0.09\columnwidth}\raggedright
137.3\strut
\end{minipage} & \begin{minipage}[t]{0.15\columnwidth}\raggedright
0.030\strut
\end{minipage}\tabularnewline
\begin{minipage}[t]{0.07\columnwidth}\raggedright
HH Pandya\strut
\end{minipage} & \begin{minipage}[t]{0.13\columnwidth}\raggedright
844\strut
\end{minipage} & \begin{minipage}[t]{0.13\columnwidth}\raggedright
512\strut
\end{minipage} & \begin{minipage}[t]{0.08\columnwidth}\raggedright
171.6\strut
\end{minipage} & \begin{minipage}[t]{0.15\columnwidth}\raggedright
0.070\strut
\end{minipage} & \begin{minipage}[t]{0.09\columnwidth}\raggedright
134.0\strut
\end{minipage} & \begin{minipage}[t]{0.15\columnwidth}\raggedright
0.037\strut
\end{minipage}\tabularnewline
\begin{minipage}[t]{0.07\columnwidth}\raggedright
Naman Dhir\strut
\end{minipage} & \begin{minipage}[t]{0.13\columnwidth}\raggedright
498\strut
\end{minipage} & \begin{minipage}[t]{0.13\columnwidth}\raggedright
204\strut
\end{minipage} & \begin{minipage}[t]{0.08\columnwidth}\raggedright
\textbf{203.9}\strut
\end{minipage} & \begin{minipage}[t]{0.15\columnwidth}\raggedright
0.070\strut
\end{minipage} & \begin{minipage}[t]{0.09\columnwidth}\raggedright
145.9\strut
\end{minipage} & \begin{minipage}[t]{0.15\columnwidth}\raggedright
0.049\strut
\end{minipage}\tabularnewline
\bottomrule
\end{longtable}

The critical structural feature is the \textbf{phase asymmetry of Naman
Dhir}: his Death SR (204) exceeds any other candidate by at least 18
points, yet his Middle SR (146) is comparable to SA Yadav's. This makes
him uniquely valuable in the Death overs but risky to promote early. SA
Yadav is the optimal bridge through the Middle overs: his dismissal
probability (0.035) is the lowest of all candidates, and his Middle-over
strike rate (143.8) is the highest among the non-Death specialists. The
combination of low wicket risk and competitive scoring rate in the
Middle phase makes him the best choice to carry the innings to the Death
overs while preserving Tilak Varma and Naman Dhir, both superior
Death-over hitters, for the final assault.

\begin{figure}
\centering
\includegraphics[width=0.75\textwidth,height=\textheight]{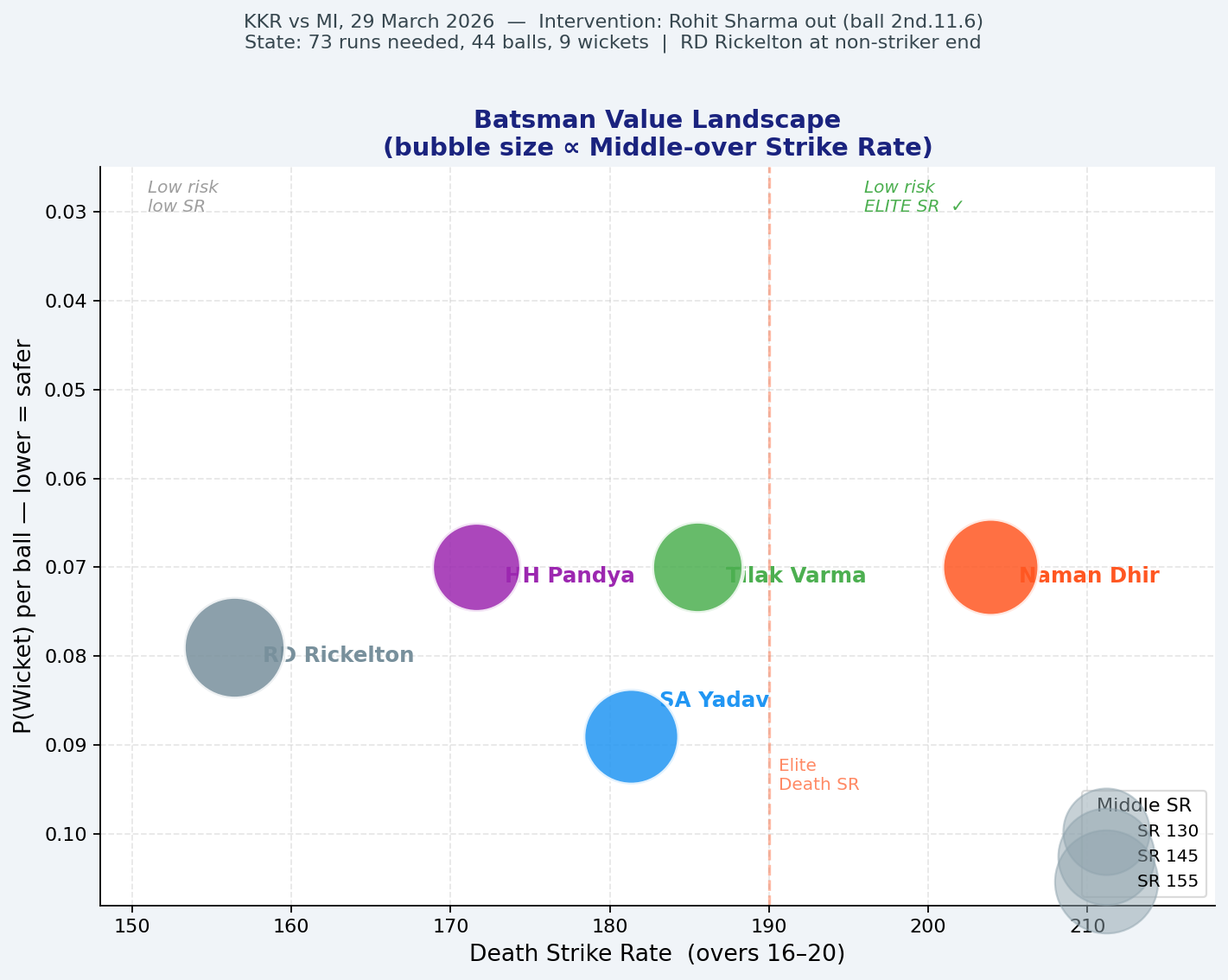}
\caption{Figure 1: MI batsman phase-specific profiles (KKR vs MI, over
12 intervention). The phase asymmetry of Naman Dhir is the key
structural feature driving the optimal batting order.}
\end{figure}

\hypertarget{batting-order-results}{%
\subsubsection{7.3 Batting Order Results}\label{batting-order-results}}

\textbf{Table 3: Best Achievable Win\% by Next-In Choice (Position 3)}

\begin{longtable}[]{@{}lll@{}}
\toprule
Next-In & Best Win\% & Optimal Remaining Order\tabularnewline
\midrule
\endhead
\textbf{SA Yadav (actual)} & \textbf{56.5\%} & Naman Dhir → Tilak Varma
→ HH Pandya\tabularnewline
Naman Dhir & 55.7\% & SA Yadav → Tilak Varma → HH Pandya\tabularnewline
Tilak Varma & 55.1\% & Naman Dhir → SA Yadav → HH Pandya\tabularnewline
HH Pandya & 53.3\% & Naman Dhir → Tilak Varma → SA Yadav\tabularnewline
\bottomrule
\end{longtable}

The actual decision to send SA Yadav in immediately at Position 3 was
the correct choice. His low Middle-over dismissal probability
(\(p^W_{\text{MI}} = 0.035\), lowest of all candidates) and competitive
Middle-over strike rate (143.8) make him the optimal bridge through
overs 12--15 while preserving Tilak Varma and Naman Dhir for the
Death-over assault.

\textbf{Table 4: All 6 Orderings for Positions 4--6 (SA Yadav at
Position 3)}

\begin{longtable}[]{@{}lllll@{}}
\toprule
Rank & Position 4 & Position 5 & Position 6 & Win\%\tabularnewline
\midrule
\endhead
\textbf{1 (opt)} & \textbf{Naman Dhir} & \textbf{Tilak Varma} &
\textbf{HH Pandya} & \textbf{56.5\%}\tabularnewline
2 & Naman Dhir & HH Pandya & Tilak Varma & 56.1\%\tabularnewline
3 & Tilak Varma & Naman Dhir & HH Pandya & 55.2\%\tabularnewline
4 & HH Pandya & Naman Dhir & Tilak Varma & 52.9\%\tabularnewline
\textbf{5 (actual)} & \textbf{Tilak Varma} & \textbf{HH Pandya} &
\textbf{Naman Dhir} & \textbf{52.4\%}\tabularnewline
6 & HH Pandya & Tilak Varma & Naman Dhir & 51.8\%\tabularnewline
\bottomrule
\end{longtable}

The match order (rank 5) was the second-worst of all six possibilities.
Figure 2 shows the best achievable win probability as a function of the
next-in choice at Position 3.

\begin{figure}
\centering
\includegraphics[width=0.6\textwidth,height=\textheight]{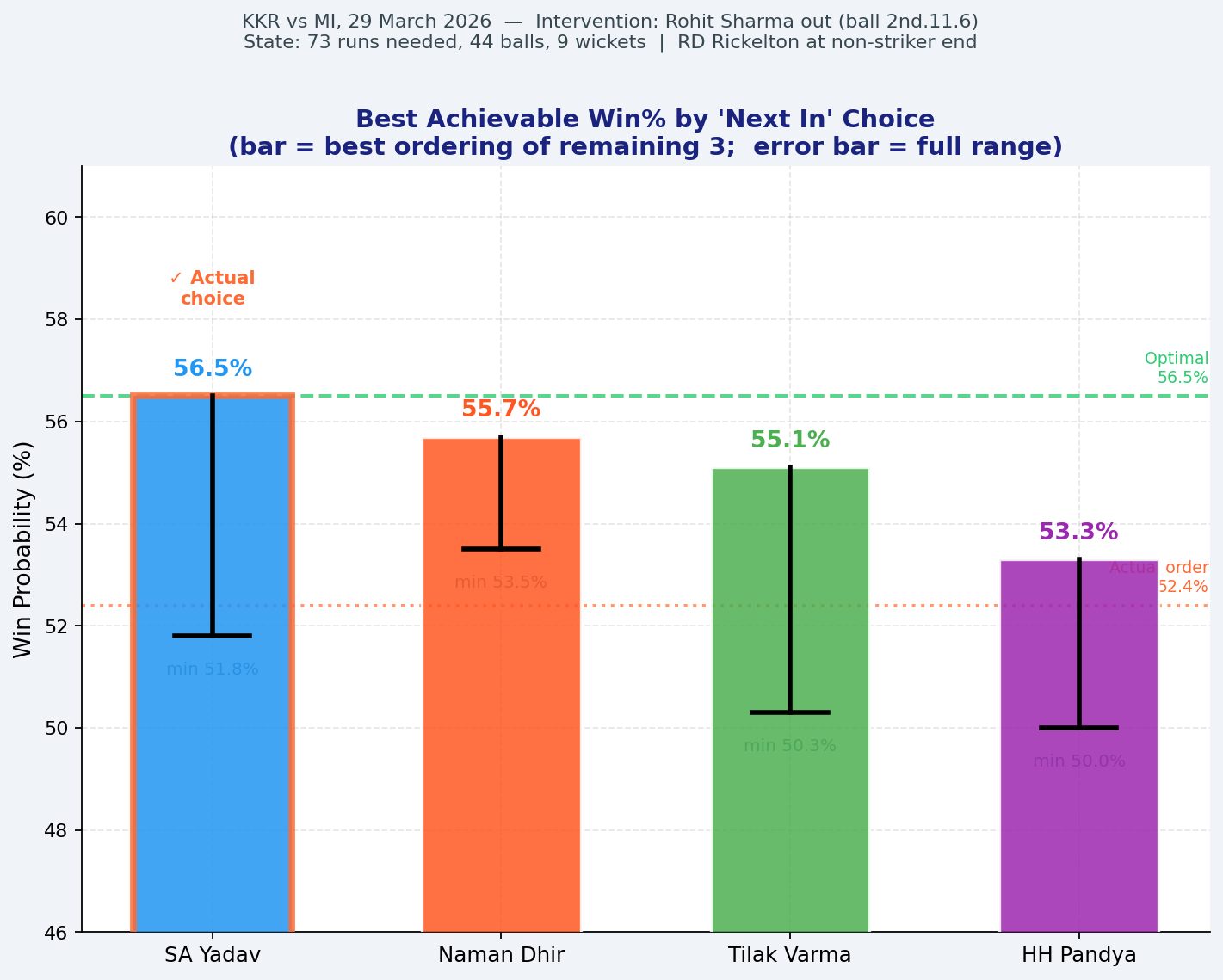}
\caption{Figure 2: Best achievable win probability by next-in choice at
Position 3. SA Yadav's low Middle-over dismissal probability makes him
the optimal bridge.}
\end{figure}

The optimal order promotes Naman Dhir to Position 4, ensuring maximum
exposure of his Death SR (204) to the final overs. In the actual match,
Dhir entered at Position 6 and faced only 2 balls, both in over 19. The
expected additional runs from promoting Dhir two positions can be
approximated as:

\[\Delta R \approx \Delta b_{\text{Dhir}} \times \frac{\text{SR}_{\text{Dhir}}^{\text{DE}} - \text{SR}_{\text{Varma}}^{\text{DE}}}{100} \approx \Delta b_{\text{Dhir}} \times 0.184\]

Even 5 additional Death balls yields approximately 0.92 expected
additional runs, sufficient given the proximity to the target, to shift
win probability by the observed 4.1 pp.

\hypertarget{summary}{%
\subsubsection{7.4 Summary}\label{summary}}

\begin{longtable}[]{@{}ll@{}}
\toprule
Scenario & Win\% (MI)\tabularnewline
\midrule
\endhead
Worst possible batting order & 50.0\%\tabularnewline
Actual batting order & 52.4\%\tabularnewline
Optimal batting order & 56.5\%\tabularnewline
Gain from optimal order & \textbf{+4.1 pp}\tabularnewline
\bottomrule
\end{longtable}

\hypertarget{case-study-2-gt-vs-pbks-31-march-2026}{%
\subsection{8. Case Study 2: GT vs PBKS, 31 March
2026}\label{case-study-2-gt-vs-pbks-31-march-2026}}

\hypertarget{match-context-and-intervention-point-1}{%
\subsubsection{8.1 Match Context and Intervention
Point}\label{match-context-and-intervention-point-1}}

Gujarat Titans posted 162 in their first innings. At ball
\texttt{2nd.9.3}, the third delivery of over 10 (0-indexed over 9),
Punjab Kings' opener P Simran Singh was caught off Rashid Khan's
bowling. At the moment of dismissal:

\begin{longtable}[]{@{}ll@{}}
\toprule
\begin{minipage}[b]{0.57\columnwidth}\raggedright
Parameter\strut
\end{minipage} & \begin{minipage}[b]{0.37\columnwidth}\raggedright
Value\strut
\end{minipage}\tabularnewline
\midrule
\endhead
\begin{minipage}[t]{0.57\columnwidth}\raggedright
PBKS score\strut
\end{minipage} & \begin{minipage}[t]{0.37\columnwidth}\raggedright
83/2\strut
\end{minipage}\tabularnewline
\begin{minipage}[t]{0.57\columnwidth}\raggedright
Runs needed (\(r_0 = d_0\))\strut
\end{minipage} & \begin{minipage}[t]{0.37\columnwidth}\raggedright
80\strut
\end{minipage}\tabularnewline
\begin{minipage}[t]{0.57\columnwidth}\raggedright
Legal balls remaining (\(b_0\))\strut
\end{minipage} & \begin{minipage}[t]{0.37\columnwidth}\raggedright
60 (overs 10--19 after Rashid completes over 9)\strut
\end{minipage}\tabularnewline
\begin{minipage}[t]{0.57\columnwidth}\raggedright
PBKS wickets in hand\strut
\end{minipage} & \begin{minipage}[t]{0.37\columnwidth}\raggedright
8\strut
\end{minipage}\tabularnewline
\begin{minipage}[t]{0.57\columnwidth}\raggedright
Required run rate\strut
\end{minipage} & \begin{minipage}[t]{0.37\columnwidth}\raggedright
8.00 per over\strut
\end{minipage}\tabularnewline
\bottomrule
\end{longtable}

Rashid Khan was mid-over at the moment of dismissal and was committed to
completing over 9 (3 remaining balls). The optimisation covers the 10
complete overs that follow (overs 10--19, 0-indexed).

\hypertarget{gt-bowling-resources-and-profiles}{%
\subsubsection{8.2 GT Bowling Resources and
Profiles}\label{gt-bowling-resources-and-profiles}}

\textbf{Table 7: GT Bowler Profiles and Quotas (After Over 9 Completes)}

\begin{longtable}[]{@{}lllllllll@{}}
\toprule
\begin{minipage}[b]{0.05\columnwidth}\raggedright
Bowler\strut
\end{minipage} & \begin{minipage}[b]{0.08\columnwidth}\raggedright
Overs bowled\strut
\end{minipage} & \begin{minipage}[b]{0.07\columnwidth}\raggedright
Quota left\strut
\end{minipage} & \begin{minipage}[b]{0.10\columnwidth}\raggedright
Hist. balls (MI)\strut
\end{minipage} & \begin{minipage}[b]{0.10\columnwidth}\raggedright
Hist. balls (DE)\strut
\end{minipage} & \begin{minipage}[b]{0.07\columnwidth}\raggedright
Middle ER\strut
\end{minipage} & \begin{minipage}[b]{0.11\columnwidth}\raggedright
\(p^W_{\text{MI}}\)\strut
\end{minipage} & \begin{minipage}[b]{0.06\columnwidth}\raggedright
Death ER\strut
\end{minipage} & \begin{minipage}[b]{0.11\columnwidth}\raggedright
\(p^W_{\text{DE}}\)\strut
\end{minipage}\tabularnewline
\midrule
\endhead
\begin{minipage}[t]{0.05\columnwidth}\raggedright
Ashok Sharma\strut
\end{minipage} & \begin{minipage}[t]{0.08\columnwidth}\raggedright
1\strut
\end{minipage} & \begin{minipage}[t]{0.07\columnwidth}\raggedright
3\strut
\end{minipage} & \begin{minipage}[t]{0.10\columnwidth}\raggedright
0\strut
\end{minipage} & \begin{minipage}[t]{0.10\columnwidth}\raggedright
0\strut
\end{minipage} & \begin{minipage}[t]{0.07\columnwidth}\raggedright
7.49†\strut
\end{minipage} & \begin{minipage}[t]{0.11\columnwidth}\raggedright
0.043†\strut
\end{minipage} & \begin{minipage}[t]{0.06\columnwidth}\raggedright
9.37†\strut
\end{minipage} & \begin{minipage}[t]{0.11\columnwidth}\raggedright
0.083†\strut
\end{minipage}\tabularnewline
\begin{minipage}[t]{0.05\columnwidth}\raggedright
K Rabada\strut
\end{minipage} & \begin{minipage}[t]{0.08\columnwidth}\raggedright
2\strut
\end{minipage} & \begin{minipage}[t]{0.07\columnwidth}\raggedright
2\strut
\end{minipage} & \begin{minipage}[t]{0.10\columnwidth}\raggedright
466\strut
\end{minipage} & \begin{minipage}[t]{0.10\columnwidth}\raggedright
599\strut
\end{minipage} & \begin{minipage}[t]{0.07\columnwidth}\raggedright
7.35\strut
\end{minipage} & \begin{minipage}[t]{0.11\columnwidth}\raggedright
0.053\strut
\end{minipage} & \begin{minipage}[t]{0.06\columnwidth}\raggedright
9.36\strut
\end{minipage} & \begin{minipage}[t]{0.11\columnwidth}\raggedright
0.104\strut
\end{minipage}\tabularnewline
\begin{minipage}[t]{0.05\columnwidth}\raggedright
Mohammed Siraj\strut
\end{minipage} & \begin{minipage}[t]{0.08\columnwidth}\raggedright
2\strut
\end{minipage} & \begin{minipage}[t]{0.07\columnwidth}\raggedright
2\strut
\end{minipage} & \begin{minipage}[t]{0.10\columnwidth}\raggedright
451\strut
\end{minipage} & \begin{minipage}[t]{0.10\columnwidth}\raggedright
673\strut
\end{minipage} & \begin{minipage}[t]{0.07\columnwidth}\raggedright
7.35\strut
\end{minipage} & \begin{minipage}[t]{0.11\columnwidth}\raggedright
0.048\strut
\end{minipage} & \begin{minipage}[t]{0.06\columnwidth}\raggedright
9.55\strut
\end{minipage} & \begin{minipage}[t]{0.11\columnwidth}\raggedright
0.067\strut
\end{minipage}\tabularnewline
\begin{minipage}[t]{0.05\columnwidth}\raggedright
Rashid Khan\strut
\end{minipage} & \begin{minipage}[t]{0.08\columnwidth}\raggedright
3*\strut
\end{minipage} & \begin{minipage}[t]{0.07\columnwidth}\raggedright
1\strut
\end{minipage} & \begin{minipage}[t]{0.10\columnwidth}\raggedright
2,223\strut
\end{minipage} & \begin{minipage}[t]{0.10\columnwidth}\raggedright
564\strut
\end{minipage} & \begin{minipage}[t]{0.07\columnwidth}\raggedright
\textbf{6.58}\strut
\end{minipage} & \begin{minipage}[t]{0.11\columnwidth}\raggedright
0.047\strut
\end{minipage} & \begin{minipage}[t]{0.06\columnwidth}\raggedright
\textbf{8.40}\strut
\end{minipage} & \begin{minipage}[t]{0.11\columnwidth}\raggedright
0.075\strut
\end{minipage}\tabularnewline
\begin{minipage}[t]{0.05\columnwidth}\raggedright
Washington Sundar\strut
\end{minipage} & \begin{minipage}[t]{0.08\columnwidth}\raggedright
2\strut
\end{minipage} & \begin{minipage}[t]{0.07\columnwidth}\raggedright
2\strut
\end{minipage} & \begin{minipage}[t]{0.10\columnwidth}\raggedright
654\strut
\end{minipage} & \begin{minipage}[t]{0.10\columnwidth}\raggedright
81\strut
\end{minipage} & \begin{minipage}[t]{0.07\columnwidth}\raggedright
7.07\strut
\end{minipage} & \begin{minipage}[t]{0.11\columnwidth}\raggedright
0.033\strut
\end{minipage} & \begin{minipage}[t]{0.06\columnwidth}\raggedright
9.61\strut
\end{minipage} & \begin{minipage}[t]{0.11\columnwidth}\raggedright
0.068\strut
\end{minipage}\tabularnewline
\begin{minipage}[t]{0.05\columnwidth}\raggedright
M Prasidh Krishna\strut
\end{minipage} & \begin{minipage}[t]{0.08\columnwidth}\raggedright
0\strut
\end{minipage} & \begin{minipage}[t]{0.07\columnwidth}\raggedright
4\strut
\end{minipage} & \begin{minipage}[t]{0.10\columnwidth}\raggedright
398\strut
\end{minipage} & \begin{minipage}[t]{0.10\columnwidth}\raggedright
505\strut
\end{minipage} & \begin{minipage}[t]{0.07\columnwidth}\raggedright
7.47\strut
\end{minipage} & \begin{minipage}[t]{0.11\columnwidth}\raggedright
0.047\strut
\end{minipage} & \begin{minipage}[t]{0.06\columnwidth}\raggedright
\textbf{9.79}\strut
\end{minipage} & \begin{minipage}[t]{0.11\columnwidth}\raggedright
0.086\strut
\end{minipage}\tabularnewline
\bottomrule
\end{longtable}

*Rashid Khan bowled 3 full overs plus 3 balls of over 9; quota = 1 full
remaining over.\\
†Ashok Sharma profile = population average (\(\lambda = 0\)); zero
historical IPL deliveries in dataset.

\begin{figure}
\centering
\includegraphics[width=0.65\textwidth,height=\textheight]{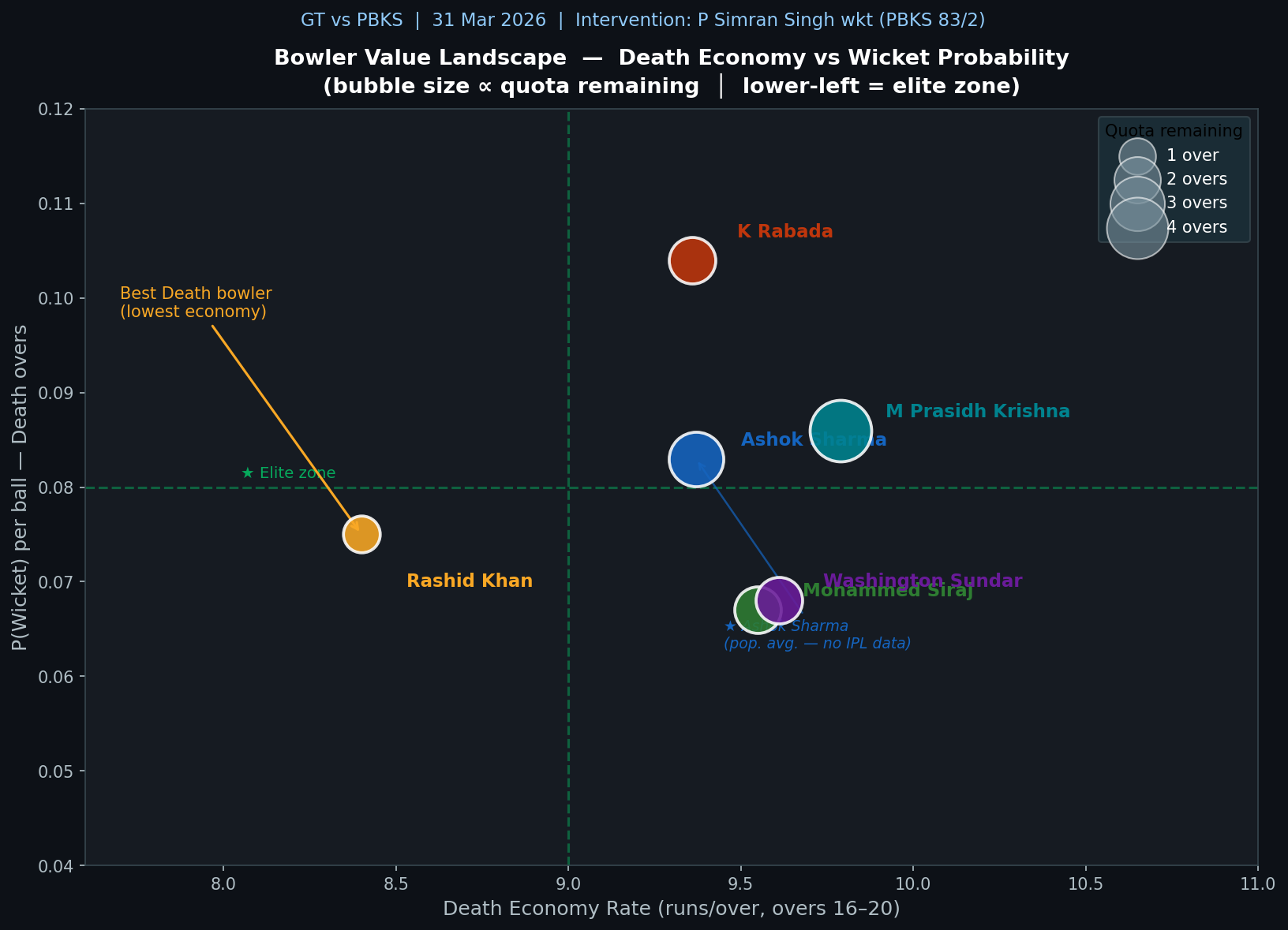}
\caption{Figure 5: GT bowler landscape at the over 10 intervention
point. The optimal Death-over bowler (Rashid Khan) is in the top-left;
the most expensive deployed bowler (M Prasidh Krishna) is toward the
bottom-right.}
\end{figure}

Key features of this profile table: - \textbf{Rashid Khan} has the
lowest economy in both Middle (6.58) and Death (8.40) phases, making him
the most valuable bowler, but his remaining quota is only 1 over. -
\textbf{Washington Sundar} shows a sharp phase gradient: Middle ER 7.07
(competitive) vs Death ER 9.61 (second-most expensive). He should not
bowl Death overs. - \textbf{M Prasidh Krishna} has the highest Death
economy (9.79) of any GT bowler, yet has 4 overs of remaining quota. -
\textbf{Mohammed Siraj} was not used in the actual plan despite 2 overs
of quota and competitive profiles in both phases (Middle ER 7.35, Death
ER 9.55).

\begin{figure}
\centering
\includegraphics[width=0.6\textwidth,height=\textheight]{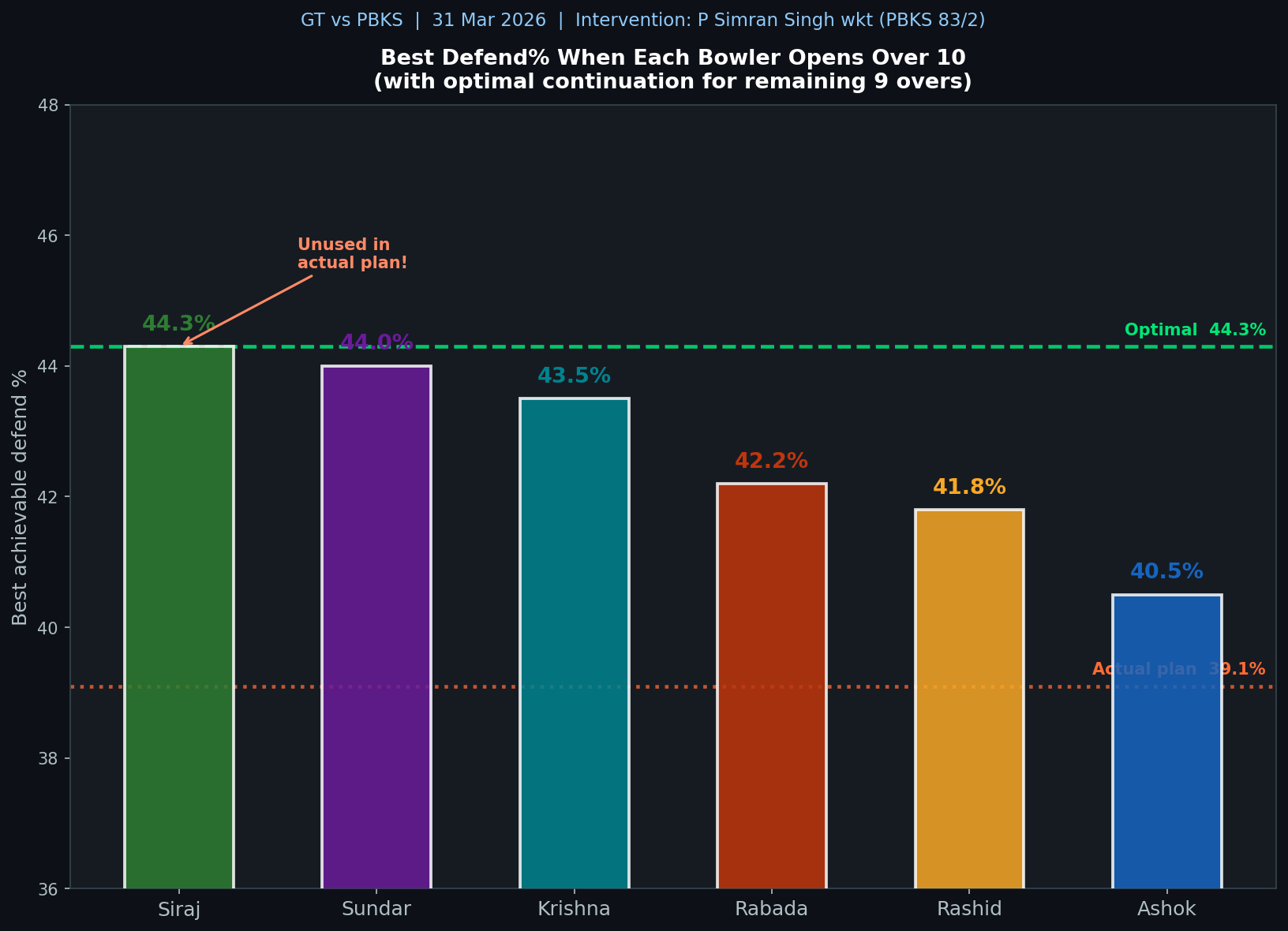}
\caption{Figure 6: Best achievable defend probability by over 10 opener
choice. Siraj's two-over quota makes him the highest-value opener
despite not being unused in the actual plan.}
\end{figure}

\hypertarget{actual-and-optimal-bowling-plans}{%
\subsubsection{8.3 Actual and Optimal Bowling
Plans}\label{actual-and-optimal-bowling-plans}}

\textbf{Table 8: Actual GT Bowling Plan (Overs 10--19)}

\begin{longtable}[]{@{}llll@{}}
\toprule
Over & Bowler & Phase & Economy\tabularnewline
\midrule
\endhead
10 & Ashok Sharma & MI & 7.49\tabularnewline
11 & Rashid Khan & MI & 6.58\tabularnewline
12 & M Prasidh Krishna & MI & 7.47\tabularnewline
13 & Washington Sundar & MI & 7.07\tabularnewline
14 & M Prasidh Krishna & MI & 7.47\tabularnewline
15 & K Rabada & DE & 9.36\tabularnewline
16 & M Prasidh Krishna & DE & 9.79\tabularnewline
17 & Ashok Sharma & DE & 9.37\tabularnewline
18 & M Prasidh Krishna & DE & 9.79\tabularnewline
19 & Washington Sundar & DE & 9.61\tabularnewline
& \textbf{Actual defend\%} & & \textbf{39.1\%}\tabularnewline
\bottomrule
\end{longtable}

\textbf{Table 9: Optimal GT Bowling Plan (Overs 10--19)}

\begin{longtable}[]{@{}lllll@{}}
\toprule
Over & Bowler & Phase & Economy & Changed?\tabularnewline
\midrule
\endhead
10 & \textbf{Mohammed Siraj} & MI & 7.35 & ←\tabularnewline
11 & \textbf{Washington Sundar} & MI & 7.07 & ←\tabularnewline
12 & \textbf{K Rabada} & MI & 7.35 & ←\tabularnewline
13 & \textbf{Mohammed Siraj} & MI & 7.35 & ←\tabularnewline
14 & \textbf{Washington Sundar} & MI & 7.07 & ←\tabularnewline
15 & \textbf{Ashok Sharma} & DE & 9.37 & ←\tabularnewline
16 & \textbf{Rashid Khan} & DE & 8.40 & ←\tabularnewline
17 & Ashok Sharma & DE & 9.37 & ---\tabularnewline
18 & \textbf{K Rabada} & DE & 9.36 & ←\tabularnewline
19 & \textbf{Ashok Sharma} & DE & 9.37 & ←\tabularnewline
& \textbf{Optimal defend\%} & & \textbf{44.3\%} &\tabularnewline
\bottomrule
\end{longtable}

\textbf{Table 10: Top-5 GT Bowling Plans}

\begin{longtable}[]{@{}llllllllllll@{}}
\toprule
\begin{minipage}[b]{0.06\columnwidth}\raggedright
Rank\strut
\end{minipage} & \begin{minipage}[b]{0.06\columnwidth}\raggedright
Ov10\strut
\end{minipage} & \begin{minipage}[b]{0.06\columnwidth}\raggedright
Ov11\strut
\end{minipage} & \begin{minipage}[b]{0.06\columnwidth}\raggedright
Ov12\strut
\end{minipage} & \begin{minipage}[b]{0.06\columnwidth}\raggedright
Ov13\strut
\end{minipage} & \begin{minipage}[b]{0.06\columnwidth}\raggedright
Ov14\strut
\end{minipage} & \begin{minipage}[b]{0.06\columnwidth}\raggedright
Ov15\strut
\end{minipage} & \begin{minipage}[b]{0.06\columnwidth}\raggedright
Ov16\strut
\end{minipage} & \begin{minipage}[b]{0.06\columnwidth}\raggedright
Ov17\strut
\end{minipage} & \begin{minipage}[b]{0.06\columnwidth}\raggedright
Ov18\strut
\end{minipage} & \begin{minipage}[b]{0.06\columnwidth}\raggedright
Ov19\strut
\end{minipage} & \begin{minipage}[b]{0.08\columnwidth}\raggedright
Defend\%\strut
\end{minipage}\tabularnewline
\midrule
\endhead
\begin{minipage}[t]{0.06\columnwidth}\raggedright
\textbf{1 (opt)}\strut
\end{minipage} & \begin{minipage}[t]{0.06\columnwidth}\raggedright
Siraj\strut
\end{minipage} & \begin{minipage}[t]{0.06\columnwidth}\raggedright
Sundar\strut
\end{minipage} & \begin{minipage}[t]{0.06\columnwidth}\raggedright
Rabada\strut
\end{minipage} & \begin{minipage}[t]{0.06\columnwidth}\raggedright
Siraj\strut
\end{minipage} & \begin{minipage}[t]{0.06\columnwidth}\raggedright
Sundar\strut
\end{minipage} & \begin{minipage}[t]{0.06\columnwidth}\raggedright
Ashok\strut
\end{minipage} & \begin{minipage}[t]{0.06\columnwidth}\raggedright
\textbf{Rashid}\strut
\end{minipage} & \begin{minipage}[t]{0.06\columnwidth}\raggedright
Ashok\strut
\end{minipage} & \begin{minipage}[t]{0.06\columnwidth}\raggedright
Rabada\strut
\end{minipage} & \begin{minipage}[t]{0.06\columnwidth}\raggedright
Ashok\strut
\end{minipage} & \begin{minipage}[t]{0.08\columnwidth}\raggedright
\textbf{44.3\%}\strut
\end{minipage}\tabularnewline
\begin{minipage}[t]{0.06\columnwidth}\raggedright
2\strut
\end{minipage} & \begin{minipage}[t]{0.06\columnwidth}\raggedright
Sundar\strut
\end{minipage} & \begin{minipage}[t]{0.06\columnwidth}\raggedright
Krishna\strut
\end{minipage} & \begin{minipage}[t]{0.06\columnwidth}\raggedright
Sundar\strut
\end{minipage} & \begin{minipage}[t]{0.06\columnwidth}\raggedright
Siraj\strut
\end{minipage} & \begin{minipage}[t]{0.06\columnwidth}\raggedright
Rabada\strut
\end{minipage} & \begin{minipage}[t]{0.06\columnwidth}\raggedright
Ashok\strut
\end{minipage} & \begin{minipage}[t]{0.06\columnwidth}\raggedright
Rashid\strut
\end{minipage} & \begin{minipage}[t]{0.06\columnwidth}\raggedright
Ashok\strut
\end{minipage} & \begin{minipage}[t]{0.06\columnwidth}\raggedright
Rabada\strut
\end{minipage} & \begin{minipage}[t]{0.06\columnwidth}\raggedright
Ashok\strut
\end{minipage} & \begin{minipage}[t]{0.08\columnwidth}\raggedright
44.0\%\strut
\end{minipage}\tabularnewline
\begin{minipage}[t]{0.06\columnwidth}\raggedright
3\strut
\end{minipage} & \begin{minipage}[t]{0.06\columnwidth}\raggedright
Sundar\strut
\end{minipage} & \begin{minipage}[t]{0.06\columnwidth}\raggedright
Siraj\strut
\end{minipage} & \begin{minipage}[t]{0.06\columnwidth}\raggedright
Sundar\strut
\end{minipage} & \begin{minipage}[t]{0.06\columnwidth}\raggedright
Rabada\strut
\end{minipage} & \begin{minipage}[t]{0.06\columnwidth}\raggedright
Siraj\strut
\end{minipage} & \begin{minipage}[t]{0.06\columnwidth}\raggedright
Ashok\strut
\end{minipage} & \begin{minipage}[t]{0.06\columnwidth}\raggedright
Rashid\strut
\end{minipage} & \begin{minipage}[t]{0.06\columnwidth}\raggedright
Ashok\strut
\end{minipage} & \begin{minipage}[t]{0.06\columnwidth}\raggedright
Rabada\strut
\end{minipage} & \begin{minipage}[t]{0.06\columnwidth}\raggedright
Ashok\strut
\end{minipage} & \begin{minipage}[t]{0.08\columnwidth}\raggedright
43.9\%\strut
\end{minipage}\tabularnewline
\begin{minipage}[t]{0.06\columnwidth}\raggedright
4\strut
\end{minipage} & \begin{minipage}[t]{0.06\columnwidth}\raggedright
Siraj\strut
\end{minipage} & \begin{minipage}[t]{0.06\columnwidth}\raggedright
Sundar\strut
\end{minipage} & \begin{minipage}[t]{0.06\columnwidth}\raggedright
Siraj\strut
\end{minipage} & \begin{minipage}[t]{0.06\columnwidth}\raggedright
Krishna\strut
\end{minipage} & \begin{minipage}[t]{0.06\columnwidth}\raggedright
Rashid\strut
\end{minipage} & \begin{minipage}[t]{0.06\columnwidth}\raggedright
Ashok\strut
\end{minipage} & \begin{minipage}[t]{0.06\columnwidth}\raggedright
Rabada\strut
\end{minipage} & \begin{minipage}[t]{0.06\columnwidth}\raggedright
Ashok\strut
\end{minipage} & \begin{minipage}[t]{0.06\columnwidth}\raggedright
Rabada\strut
\end{minipage} & \begin{minipage}[t]{0.06\columnwidth}\raggedright
Ashok\strut
\end{minipage} & \begin{minipage}[t]{0.08\columnwidth}\raggedright
43.5\%\strut
\end{minipage}\tabularnewline
\begin{minipage}[t]{0.06\columnwidth}\raggedright
\textbf{Actual}\strut
\end{minipage} & \begin{minipage}[t]{0.06\columnwidth}\raggedright
Ashok\strut
\end{minipage} & \begin{minipage}[t]{0.06\columnwidth}\raggedright
\textbf{Rashid}\strut
\end{minipage} & \begin{minipage}[t]{0.06\columnwidth}\raggedright
Krishna\strut
\end{minipage} & \begin{minipage}[t]{0.06\columnwidth}\raggedright
Sundar\strut
\end{minipage} & \begin{minipage}[t]{0.06\columnwidth}\raggedright
Krishna\strut
\end{minipage} & \begin{minipage}[t]{0.06\columnwidth}\raggedright
Rabada\strut
\end{minipage} & \begin{minipage}[t]{0.06\columnwidth}\raggedright
\textbf{Krishna}\strut
\end{minipage} & \begin{minipage}[t]{0.06\columnwidth}\raggedright
Ashok\strut
\end{minipage} & \begin{minipage}[t]{0.06\columnwidth}\raggedright
\textbf{Krishna}\strut
\end{minipage} & \begin{minipage}[t]{0.06\columnwidth}\raggedright
Sundar\strut
\end{minipage} & \begin{minipage}[t]{0.08\columnwidth}\raggedright
\textbf{39.1\%}\strut
\end{minipage}\tabularnewline
\bottomrule
\end{longtable}

\begin{figure}
\centering
\includegraphics[width=0.85\textwidth,height=\textheight]{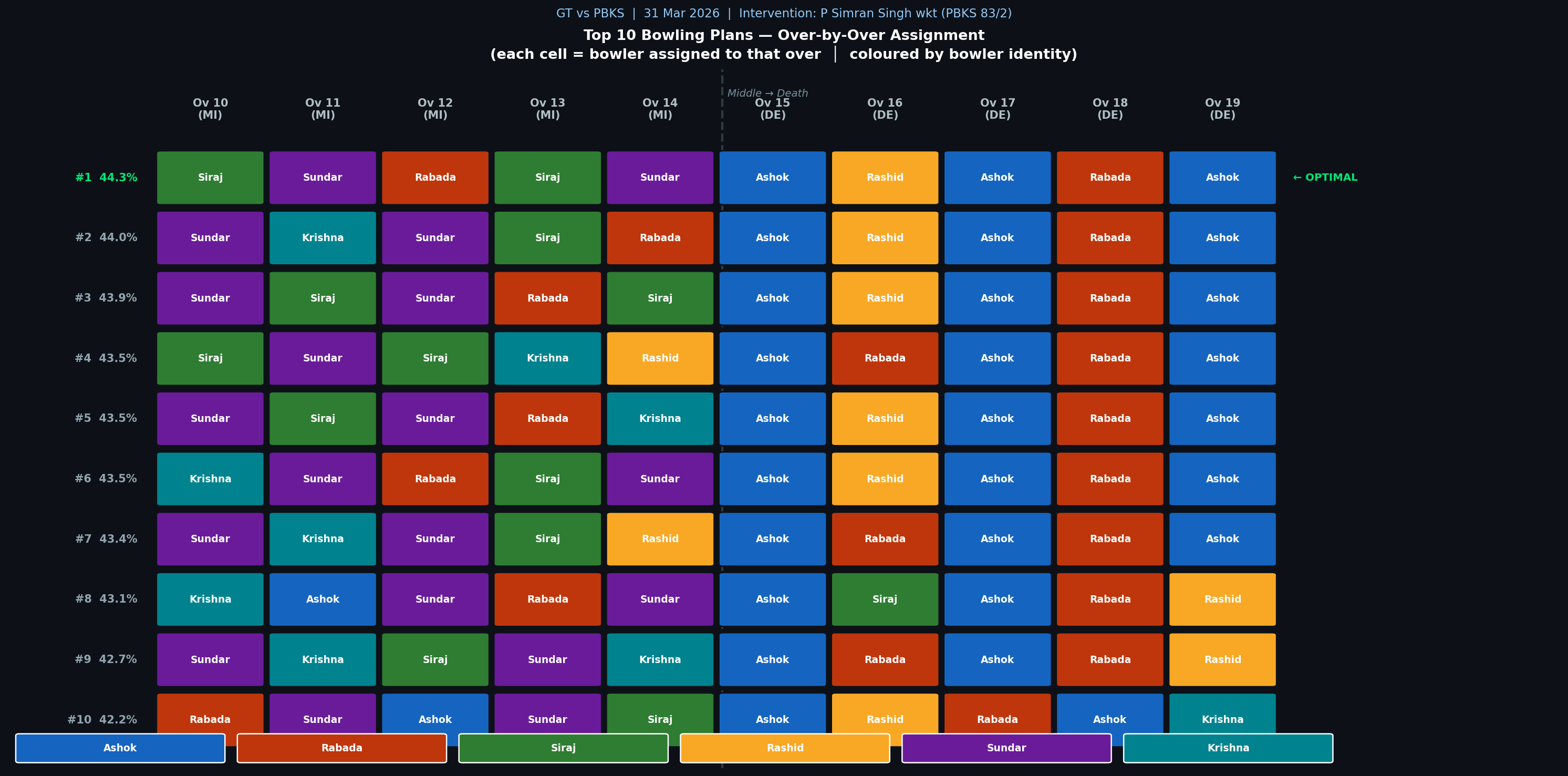}
\caption{Figure 7: Top 10 GT bowling plans heatmap. The vertical divider
separates Middle overs (left) from Death overs (right); Rashid Khan
appears in Death over 16 in all top plans.}
\end{figure}

Three structural errors characterise the actual plan:

\begin{enumerate}
\def\labelenumi{\arabic{enumi}.}
\item
  \textbf{Rashid Khan deployed in Middle (over 11)}: His advantage over
  the best alternative (Washington Sundar, ER 7.07) in Middle is 0.49
  RPO. His advantage over M Prasidh Krishna (ER 9.79) in Death is 1.39
  RPO, nearly three times as large. Deploying Rashid in Middle over 11
  rather than Death over 16 forfeited this asymmetry.
\item
  \textbf{M Prasidh Krishna used for 4 overs, including 2 Death overs
  (16, 18)}: Krishna's Death ER (9.79) is the highest of any GT bowler.
  His 2 Death overs are expected to concede approximately 1.4 additional
  runs each relative to the optimal assignment, approximately 2.8 runs
  cumulatively.
\item
  \textbf{Mohammed Siraj not used}: With 2 overs of quota and
  competitive Middle (7.35) and Death (9.55) profiles, Siraj's omission
  was the largest single resource-allocation error. The optimal plan
  uses him in overs 10 and 13 (both Middle phase).
\end{enumerate}

\begin{figure}
\centering
\includegraphics[width=0.85\textwidth,height=\textheight]{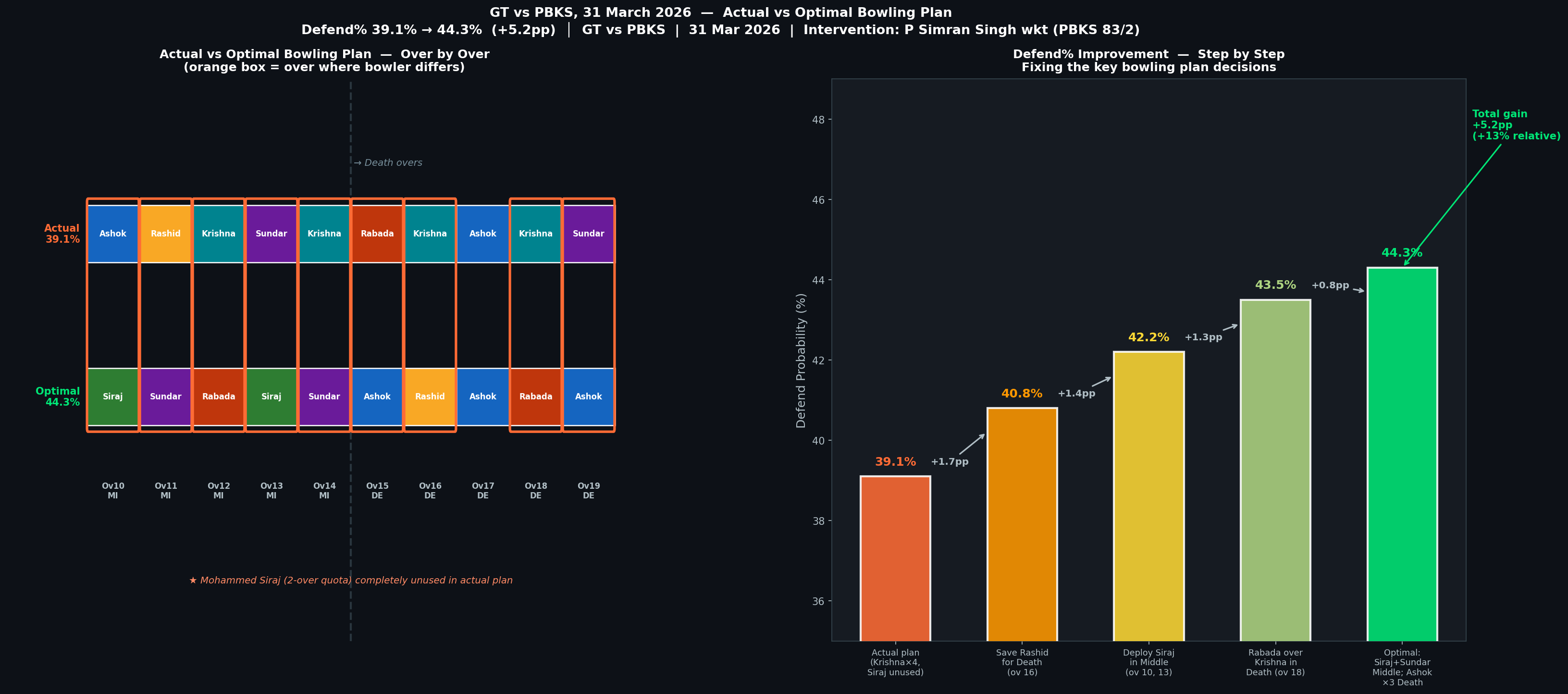}
\caption{Figure 8: Actual vs optimal GT bowling plan (overs 10--19). The
total gain of +5.2 pp is distributed across multiple substitutions, with
Siraj's inclusion and Rashid's repositioning contributing the most.}
\end{figure}

\hypertarget{statistical-significance}{%
\subsubsection{8.4 Statistical
Significance}\label{statistical-significance}}

The \(z\)-score for the bowling gap in the GT vs PBKS case:

\[z = \frac{0.052}{\sqrt{2} \times 0.00187} \approx 19.7\]

This is far beyond any conventional significance threshold, confirming
that the observed 5.2 pp gap is a stable structural property of the
plan, not simulation variance.

\hypertarget{comparative-analysis-and-discussion}{%
\subsection{9. Comparative Analysis and
Discussion}\label{comparative-analysis-and-discussion}}

\hypertarget{the-phase-specificity-of-bowling-value}{%
\subsubsection{9.1 The Phase-Specificity of Bowling
Value}\label{the-phase-specificity-of-bowling-value}}

The GT vs PBKS case illustrates the core finding: \textbf{aggregate
economy rate is a poor proxy for in-match bowling value}. M Prasidh
Krishna has an acceptable overall economy, yet his Death-over economy
(9.79 RPO) is the highest of any GT bowler. This divergence is invisible
in aggregate statistics but is fully captured by the phase-specific
profile. The primary misallocations in the GT plan are summarised in
Table 11.

\textbf{Table 11: Primary Misallocations --- GT vs PBKS}

\begin{longtable}[]{@{}llllll@{}}
\toprule
\begin{minipage}[b]{0.15\columnwidth}\raggedright
Bowler misused\strut
\end{minipage} & \begin{minipage}[b]{0.16\columnwidth}\raggedright
Phase deployed\strut
\end{minipage} & \begin{minipage}[b]{0.10\columnwidth}\raggedright
Death ER\strut
\end{minipage} & \begin{minipage}[b]{0.15\columnwidth}\raggedright
Better option\strut
\end{minipage} & \begin{minipage}[b]{0.17\columnwidth}\raggedright
Better Death ER\strut
\end{minipage} & \begin{minipage}[b]{0.12\columnwidth}\raggedright
Cost (RPO)\strut
\end{minipage}\tabularnewline
\midrule
\endhead
\begin{minipage}[t]{0.15\columnwidth}\raggedright
MP Krishna\strut
\end{minipage} & \begin{minipage}[t]{0.16\columnwidth}\raggedright
Death (2 overs)\strut
\end{minipage} & \begin{minipage}[t]{0.10\columnwidth}\raggedright
9.79\strut
\end{minipage} & \begin{minipage}[t]{0.15\columnwidth}\raggedright
Rashid Khan\strut
\end{minipage} & \begin{minipage}[t]{0.17\columnwidth}\raggedright
8.40\strut
\end{minipage} & \begin{minipage}[t]{0.12\columnwidth}\raggedright
−1.39\strut
\end{minipage}\tabularnewline
\begin{minipage}[t]{0.15\columnwidth}\raggedright
Siraj\strut
\end{minipage} & \begin{minipage}[t]{0.16\columnwidth}\raggedright
Not used\strut
\end{minipage} & \begin{minipage}[t]{0.10\columnwidth}\raggedright
---\strut
\end{minipage} & \begin{minipage}[t]{0.15\columnwidth}\raggedright
---\strut
\end{minipage} & \begin{minipage}[t]{0.17\columnwidth}\raggedright
---\strut
\end{minipage} & \begin{minipage}[t]{0.12\columnwidth}\raggedright
−0.49 per over\strut
\end{minipage}\tabularnewline
\bottomrule
\end{longtable}

Teams should maintain and consult phase-specific profiles, not aggregate
metrics, when making in-match bowling decisions. The difference between
using M Prasidh Krishna and Rashid Khan in a Death over is 1.39 expected
runs --- across two overs that is nearly 3 runs, decisive at a margin of
80 to defend in 10 overs.

\hypertarget{the-rashid-khan-problem-scarcity-of-the-best-resource}{%
\subsubsection{9.3 The Rashid Khan Problem: Scarcity of the Best
Resource}\label{the-rashid-khan-problem-scarcity-of-the-best-resource}}

The GT vs PBKS case illustrates a generalised decision problem absent
from the KKR case: \textbf{optimal allocation of a scarce high-value
resource}. Rashid Khan was the best available bowler in both Middle and
Death phases, but with only 1 over of quota remaining, the batting
captain had to choose where to deploy him. The model's recommendation,
namely Death over 16, reflects the principle that scarce resources
should be deployed where their \textbf{marginal contribution} is
largest, not merely where they are good. Since Rashid's margin over
alternatives is 1.39 RPO in Death versus 0.49 RPO in Middle, the Death
deployment is nearly three times more valuable per over. This principle
generalises to any team holding a single over of a premium bowler at the
death stage of an innings.

\hypertarget{batting-and-bowling-duality-in-practice}{%
\subsubsection{9.4 Batting and Bowling Duality in
Practice}\label{batting-and-bowling-duality-in-practice}}

The KKR vs MI case provides an audit of the batting decision at the same
intervention state where the bowling decision could also have been
studied. The batting MDP value function \(V^B\) and the bowling defend
probability \(V^\pi\) are exact complements at any shared state:
\(V^B + V^\pi = 1\). The actual batting order (rank 5 of 6) gave MI a
win probability of 52.4\%; the optimal order raises this to 56.5\%. The
same duality means that a bowling optimisation at that state would
directly reduce MI's win probability by an equivalent margin,
illustrating how both sides of the decision can be audited from the same
framework.

\hypertarget{limitations}{%
\subsubsection{9.5 Limitations}\label{limitations}}

\textbf{Static historical profiles.} Player profiles are estimated from
all historical IPL deliveries with equal weight. Current form, fatigue,
pitch conditions, and opponent-specific tendencies are not captured. A
bowler who has been expensive in recent matches, or a batsman in
exceptional form, may be systematically misrated. Exponential decay
weighting over recent innings, or Bayesian updating within the current
match, would partially address this.

\textbf{Opponent-agnostic profiles.} Batsman profiles are averaged
across all bowlers faced; bowler profiles are averaged across all
batsmen faced. Matchup-specific effects, such as a left-arm spinner
facing left-handed versus right-handed batsmen or a pace bowler against
an aggressive versus defensive batsman, are real but are not modelled
here due to the sparsity of matchup-level data across phases. A tensor
decomposition of outcomes over (bowler, batsman, phase) would extend the
framework to capture these effects.

\textbf{Pre-committed plans.} The bowling plan \(\pi\) is assumed to be
chosen once at the intervention point and then executed regardless of
how the match state evolves. In practice, the bowling captain can and
should revise the plan at each over boundary given new information (runs
conceded in the previous over, a wicket taken, a batsman promoted). A
fully adaptive policy \(\pi: (d, b, w) \to j\) would require solving the
full MDP over \((d, b, w, \mathcal{Q})\), the augmented state that
includes remaining quotas, which is feasible in principle but increases
the state space by a factor of \(\prod_j (q_j + 1)\).

\textbf{Ball-by-ball independence.} Consecutive ball outcomes are
treated as i.i.d. given the current phase and player identities.
Momentum effects such as a batsman in rhythm after a boundary or a
bowler tightening their line under pressure are not captured.
Incorporating a latent pressure or momentum state via a Hidden Markov
Model would extend the framework at the cost of additional estimation
complexity.

\hypertarget{conclusions}{%
\subsection{10. Conclusions}\label{conclusions}}

This paper has presented a unified Markov Decision Process framework for
optimising two of the most consequential in-match decisions in T20
cricket: the batting order at each wicket fall and the bowling plan
across the remaining over slots. The framework is grounded in a
three-phase player profile engine with James--Stein shrinkage for
data-sparse players, evaluated by vectorised Monte Carlo simulation, and
searched by exhaustive enumeration (batting order) and simulated
annealing (bowling plan).

Applied to two intervention points from the 2026 IPL season, the main
findings are:

\begin{enumerate}
\def\labelenumi{\arabic{enumi}.}
\item
  \textbf{Batting order (KKR vs MI):} The optimal batting order improves
  MI's win probability by 4.1 pp (52.4\% → 56.5\%). The actual order was
  the second-worst of all six feasible permutations. The primary driver
  is the phase asymmetry of Naman Dhir: his Death SR (204) is superior
  to any alternative, but he entered at Position 6 and faced only 2
  balls. Promoting him to Position 4 ensures exposure of this advantage
  to the decisive Death-over deliveries.
\item
  \textbf{Bowling plan (GT vs PBKS):} The optimal bowling plan improves
  GT's defend probability by 5.2 pp (39.1\% → 44.3\%). M Prasidh Krishna
  (Death ER = 9.79) was used in 4 overs including 2 Death overs,
  Mohammed Siraj (2-over quota, competitive profile) was not used at
  all, and Rashid Khan (best Death bowler, ER = 8.40) was deployed in a
  Middle over where his margin over alternatives was less than one-third
  of his Death margin.
\item
  \textbf{Phase-specificity as the common thread:} In both case studies,
  the source of suboptimality is phase-agnostic deployment: decisions
  that appear reasonable by aggregate metrics but are exposed as costly
  when phase-specific profiles are applied. Published aggregate
  statistics (overall SR, overall economy) suppress the information that
  matters most for in-match decision support.
\item
  \textbf{Methodological:} The simulation-based MDP approach is
  computationally tractable on commodity hardware (under 5 minutes for a
  full SA run including profile estimation), suggesting feasibility for
  pitch-side or dugout deployment during a match. With Monte Carlo SE
  below 0.22\% at \(N = 50{,}000\), the precision is sufficient to rank
  orderings or plans separated by 1 pp with high confidence.
\end{enumerate}

Phase-specific quantitative decision support, grounded in a principled
stochastic optimisation framework, offers a demonstrably large and
actionable improvement over the heuristic approach that currently
dominates in-match tactical decision-making.

\hypertarget{acknowledgements}{%
\subsection{Acknowledgements}\label{acknowledgements}}

Ball-by-ball data sourced from Cricsheet.org (Stevenson, 2023) under
Creative Commons Attribution licence. Analysis conducted in Python 3.11
using NumPy 1.26, pandas 2.1, and matplotlib 3.8. Code and figures are
available at: https://github.com/tvganesh/T20-MDP-optimisation.

\hypertarget{references}{%
\subsection{References}\label{references}}

Bellman, R. (1957). \emph{Dynamic Programming}. Princeton University
Press, Princeton, NJ.

Duckworth, F. C. and Lewis, A. J. (1998). A fair method for resetting
the target in interrupted one-day cricket matches. \emph{Journal of the
Operational Research Society}, 49(3), 220--227.

Ganesh, T. V. (2016). yorkr: An R package for analytics of cricket. R
package version 0.0.5. Available at CRAN.

Ganesh, T. V. (2017). Using linear programming for optimizing bowling
change or batting lineup in T20 cricket. \emph{gigadom.in}, 28 September
2017.

James, W. and Stein, C. (1961). Estimation with quadratic loss. In
\emph{Proceedings of the 4th Berkeley Symposium on Mathematical
Statistics and Probability}, Vol. 1, pp.~361--379. University of
California Press.

Kirkpatrick, S., Gelatt, C. D., and Vecchi, M. P. (1983). Optimization
by simulated annealing. \emph{Science}, 220(4598), 671--680.

Puterman, M. L. (1994). \emph{Markov Decision Processes: Discrete
Stochastic Dynamic Programming}. Wiley, New York.

Stevenson, S. (2023). Cricsheet: Ball-by-ball cricket data. Available
at: https://cricsheet.org. Accessed April 2026.

\hypertarget{appendix-notation-summary}{%
\subsection{Appendix: Notation
Summary}\label{appendix-notation-summary}}

\begin{longtable}[]{@{}ll@{}}
\toprule
\begin{minipage}[b]{0.38\columnwidth}\raggedright
Symbol\strut
\end{minipage} & \begin{minipage}[b]{0.56\columnwidth}\raggedright
Definition\strut
\end{minipage}\tabularnewline
\midrule
\endhead
\begin{minipage}[t]{0.38\columnwidth}\raggedright
\((r, b, w)\)\strut
\end{minipage} & \begin{minipage}[t]{0.56\columnwidth}\raggedright
Match state: runs remaining, balls remaining, wickets\strut
\end{minipage}\tabularnewline
\begin{minipage}[t]{0.38\columnwidth}\raggedright
\(R_{\max}\)\strut
\end{minipage} & \begin{minipage}[t]{0.56\columnwidth}\raggedright
Maximum runs in the state space\strut
\end{minipage}\tabularnewline
\begin{minipage}[t]{0.38\columnwidth}\raggedright
\(B = 120\)\strut
\end{minipage} & \begin{minipage}[t]{0.56\columnwidth}\raggedright
Total balls in T20 innings\strut
\end{minipage}\tabularnewline
\begin{minipage}[t]{0.38\columnwidth}\raggedright
\(\mathcal{P}\)\strut
\end{minipage} & \begin{minipage}[t]{0.56\columnwidth}\raggedright
Pool of available batsmen\strut
\end{minipage}\tabularnewline
\begin{minipage}[t]{0.38\columnwidth}\raggedright
\(\sigma\)\strut
\end{minipage} & \begin{minipage}[t]{0.56\columnwidth}\raggedright
Batting order --- permutation of \(\mathcal{P}\)\strut
\end{minipage}\tabularnewline
\begin{minipage}[t]{0.38\columnwidth}\raggedright
\(\pi = (\pi_1, \ldots, \pi_m)\)\strut
\end{minipage} & \begin{minipage}[t]{0.56\columnwidth}\raggedright
Bowling plan --- ordered over assignments\strut
\end{minipage}\tabularnewline
\begin{minipage}[t]{0.38\columnwidth}\raggedright
\(\sigma^*,\; \pi^*\)\strut
\end{minipage} & \begin{minipage}[t]{0.56\columnwidth}\raggedright
Optimal batting order, optimal bowling plan\strut
\end{minipage}\tabularnewline
\begin{minipage}[t]{0.38\columnwidth}\raggedright
\(\phi \in \{\text{PP, MI, DE}\}\)\strut
\end{minipage} & \begin{minipage}[t]{0.56\columnwidth}\raggedright
Innings phase\strut
\end{minipage}\tabularnewline
\begin{minipage}[t]{0.38\columnwidth}\raggedright
\(\Omega = \{W, 0, 1, 2, 3, 4, 6\}\)\strut
\end{minipage} & \begin{minipage}[t]{0.56\columnwidth}\raggedright
Ball outcome set\strut
\end{minipage}\tabularnewline
\begin{minipage}[t]{0.38\columnwidth}\raggedright
\(\rho(o)\)\strut
\end{minipage} & \begin{minipage}[t]{0.56\columnwidth}\raggedright
Runs scored under outcome \(o\)\strut
\end{minipage}\tabularnewline
\begin{minipage}[t]{0.38\columnwidth}\raggedright
\(\mathbf{p}_i^\phi = (p_i^{o,\phi})_{o \in \Omega}\)\strut
\end{minipage} & \begin{minipage}[t]{0.56\columnwidth}\raggedright
Outcome probability vector: player \(i\), phase \(\phi\)\strut
\end{minipage}\tabularnewline
\begin{minipage}[t]{0.38\columnwidth}\raggedright
\(T(\mathcal{S}, o)\)\strut
\end{minipage} & \begin{minipage}[t]{0.56\columnwidth}\raggedright
State transition function\strut
\end{minipage}\tabularnewline
\begin{minipage}[t]{0.38\columnwidth}\raggedright
\(V_t^B(r,b,w)\)\strut
\end{minipage} & \begin{minipage}[t]{0.56\columnwidth}\raggedright
Batting win probability at time \(t = b\) balls remaining\strut
\end{minipage}\tabularnewline
\begin{minipage}[t]{0.38\columnwidth}\raggedright
\(V_t^\pi(d,b,w)\)\strut
\end{minipage} & \begin{minipage}[t]{0.56\columnwidth}\raggedright
Bowling defend probability at time \(t = b\) balls remaining\strut
\end{minipage}\tabularnewline
\begin{minipage}[t]{0.38\columnwidth}\raggedright
\(\text{SR}_i^\phi\)\strut
\end{minipage} & \begin{minipage}[t]{0.56\columnwidth}\raggedright
Batsman \(i\) strike rate in phase \(\phi\)\strut
\end{minipage}\tabularnewline
\begin{minipage}[t]{0.38\columnwidth}\raggedright
\(\text{ER}_j^\phi\)\strut
\end{minipage} & \begin{minipage}[t]{0.56\columnwidth}\raggedright
Bowler \(j\) economy rate in phase \(\phi\)\strut
\end{minipage}\tabularnewline
\begin{minipage}[t]{0.38\columnwidth}\raggedright
\(p_i^{W,\phi}\)\strut
\end{minipage} & \begin{minipage}[t]{0.56\columnwidth}\raggedright
Dismissal/wicket probability per ball\strut
\end{minipage}\tabularnewline
\begin{minipage}[t]{0.38\columnwidth}\raggedright
\(p_j^{0,\phi}\)\strut
\end{minipage} & \begin{minipage}[t]{0.56\columnwidth}\raggedright
Dot-ball probability per ball\strut
\end{minipage}\tabularnewline
\begin{minipage}[t]{0.38\columnwidth}\raggedright
\(c_i^{o,\phi}\)\strut
\end{minipage} & \begin{minipage}[t]{0.56\columnwidth}\raggedright
Raw outcome count for player \(i\), outcome \(o\), phase \(\phi\)\strut
\end{minipage}\tabularnewline
\begin{minipage}[t]{0.38\columnwidth}\raggedright
\(\alpha = 1\)\strut
\end{minipage} & \begin{minipage}[t]{0.56\columnwidth}\raggedright
Laplace smoothing constant\strut
\end{minipage}\tabularnewline
\begin{minipage}[t]{0.38\columnwidth}\raggedright
\(\bar{\mathbf{p}}^\phi\)\strut
\end{minipage} & \begin{minipage}[t]{0.56\columnwidth}\raggedright
Population-average profile for phase \(\phi\)\strut
\end{minipage}\tabularnewline
\begin{minipage}[t]{0.38\columnwidth}\raggedright
\(\lambda_i^\phi\)\strut
\end{minipage} & \begin{minipage}[t]{0.56\columnwidth}\raggedright
James--Stein blend weight\strut
\end{minipage}\tabularnewline
\begin{minipage}[t]{0.38\columnwidth}\raggedright
\(n_i^\phi\)\strut
\end{minipage} & \begin{minipage}[t]{0.56\columnwidth}\raggedright
Historical legal deliveries for player \(i\) in phase \(\phi\)\strut
\end{minipage}\tabularnewline
\begin{minipage}[t]{0.38\columnwidth}\raggedright
\(n_{\min} = 50\)\strut
\end{minipage} & \begin{minipage}[t]{0.56\columnwidth}\raggedright
Shrinkage threshold\strut
\end{minipage}\tabularnewline
\begin{minipage}[t]{0.38\columnwidth}\raggedright
\(N = 50{,}000\)\strut
\end{minipage} & \begin{minipage}[t]{0.56\columnwidth}\raggedright
Monte Carlo simulation count (final evaluation)\strut
\end{minipage}\tabularnewline
\begin{minipage}[t]{0.38\columnwidth}\raggedright
\(N_{\text{fast}} = 5{,}000\)\strut
\end{minipage} & \begin{minipage}[t]{0.56\columnwidth}\raggedright
Monte Carlo count during SA search\strut
\end{minipage}\tabularnewline
\begin{minipage}[t]{0.38\columnwidth}\raggedright
\(\hat{V}\)\strut
\end{minipage} & \begin{minipage}[t]{0.56\columnwidth}\raggedright
Monte Carlo estimate of value function\strut
\end{minipage}\tabularnewline
\begin{minipage}[t]{0.38\columnwidth}\raggedright
\(\text{SE}(\hat{V})\)\strut
\end{minipage} & \begin{minipage}[t]{0.56\columnwidth}\raggedright
Binomial standard error of estimate\strut
\end{minipage}\tabularnewline
\begin{minipage}[t]{0.38\columnwidth}\raggedright
\(q_j\)\strut
\end{minipage} & \begin{minipage}[t]{0.56\columnwidth}\raggedright
Remaining over quota for bowler \(j\)\strut
\end{minipage}\tabularnewline
\begin{minipage}[t]{0.38\columnwidth}\raggedright
\(\mathcal{F}\)\strut
\end{minipage} & \begin{minipage}[t]{0.56\columnwidth}\raggedright
Set of feasible bowling plans\strut
\end{minipage}\tabularnewline
\begin{minipage}[t]{0.38\columnwidth}\raggedright
\(\mathcal{C}_k(\pi)\)\strut
\end{minipage} & \begin{minipage}[t]{0.56\columnwidth}\raggedright
Feasible candidate set for over slot \(k\)\strut
\end{minipage}\tabularnewline
\begin{minipage}[t]{0.38\columnwidth}\raggedright
\(A(\Delta V, T)\)\strut
\end{minipage} & \begin{minipage}[t]{0.56\columnwidth}\raggedright
SA acceptance probability\strut
\end{minipage}\tabularnewline
\begin{minipage}[t]{0.38\columnwidth}\raggedright
\(T_0 = 0.05\)\strut
\end{minipage} & \begin{minipage}[t]{0.56\columnwidth}\raggedright
Initial SA temperature\strut
\end{minipage}\tabularnewline
\begin{minipage}[t]{0.38\columnwidth}\raggedright
\(N_{\text{steps}} = 8{,}000\)\strut
\end{minipage} & \begin{minipage}[t]{0.56\columnwidth}\raggedright
SA iteration count\strut
\end{minipage}\tabularnewline
\bottomrule
\end{longtable}

\end{document}